\documentclass[nohyperref]{article}

\usepackage{microtype}
\usepackage{graphicx}
\usepackage{subfig}
\usepackage{booktabs} 

\usepackage[hyphens]{url}
\usepackage{hyperref}  

\usepackage[accepted]{icml2022}

\usepackage{amsmath}
\usepackage{amssymb}
\usepackage{mathtools}
\usepackage{amsthm}

\usepackage[capitalize,noabbrev]{cleveref}

\theoremstyle{plain}
\newtheorem{theorem}{Theorem}[section]

\newtheorem{lemma}[theorem]{Lemma}
\newtheorem{corollary}[theorem]{Corollary}
\theoremstyle{definition}

\theoremstyle{remark}

\usepackage[textsize=tiny]{todonotes}

\icmltitlerunning{\name: Communication-Efficient and Robust
Distributed Mean Estimation for Federated Learning}

\usepackage{stmaryrd}
\usepackage{thmtools} 
\usepackage{xspace} 
\usepackage{thm-restate}
\usepackage[shortlabels]{enumitem}
\usepackage{bbm}
\usepackage{wrapfig}
\newcommand{\T}[1]{\noindent\textbf{#1}}
\newcommand{\s}{S}
\newcommand{\name}{EDEN\xspace}
\newcommand\myeq{\stackrel{\mathclap{\normalfont\mbox{\small d}}}{=}}
\newcommand{\norm}[1]{\left\lVert#1\right\rVert}
\newcommand{\E}{\mathbb{E}}
\newcommand{\floor}[1]{\left\lfloor#1\right\rfloor}
\newcommand{\ceil}[1]{\left\lceil#1\right\rceil}
\newcommand{\parentheses}[1]{\left(#1\right)}
\newcommand{\angles}[1]{\left\langle#1\right\rangle}
\newcommand{\brackets}[1]{\left[#1\right]}
\newcommand{\set}[1]{\left\{#1\right\}}
\newcommand{\abs}[1]{\left|#1\right|}

\newcommand{\sbrac}[1]{\left[ #1 \right]}
\newcommand{\para}[1]{\left( #1 \right)} 

\newif\ifcomm
\commfalse
\ifcomm
\newcommand{\SV}[1]{\textcolor{blue}{SV:~#1}}
\newcommand{\ran}[1]{\textcolor{red}{Ran:~#1}} 
\newcommand{\YBI}[1]{\textcolor{green}{YBI:~#1}} 
\newcommand{\MM}[1]{\textcolor{purple}{MM:~#1}} 
\newcommand{\AP}[1]{\textcolor{violet}{AP:~#1}} 
\else
\newcommand{\SV}[1]{}
\newcommand{\ran}[1]{}
\newcommand{\YBI}[1]{}
\newcommand{\MM}[1]{}
\newcommand{\AP}[1]{}
\fi

\begin{document}

\twocolumn[
\icmltitle{\name: Communication-Efficient and Robust
Distributed\\ Mean Estimation for Federated Learning}



\icmlsetsymbol{equal}{*}

\author{%
Shay Vargaftik \thanks{Equal Contribution.\\} \ 
\\
VMware Research \\
 \texttt{shayv@vmware.com}
\And
Ran Ben Basat \samethanks[1]\\
University College London\\
\texttt{r.benbasat@cs.ucl.ac.uk}
\And
Amit Portnoy \samethanks[1]\\
Ben-Gurion University \\
\texttt{amitport@post.bgu.ac.il}
\And
Gal Mendelson \\
Stanford University\\
\texttt{galmen@stanford.edu}
\And
Yaniv Ben-Itzhak \\
VMware Research \\
\texttt{ybenitzhak@vmware.com}
\And
Michael Mitzenmacher \\
Harvard University \\
\texttt{michaelm@eecs.harvard.edu}
}

\begin{icmlauthorlist}
\icmlauthor{Shay Vargaftik}{equal,yyy}
\icmlauthor{Ran Ben Basat}{equal,xxx}
\icmlauthor{Amit Portnoy}{equal,zzz}\\
\icmlauthor{Gal Mendelson}{aaa}
\icmlauthor{Yaniv Ben-Itzhak}{yyy}
\icmlauthor{Michael Mitzenmacher}{bbb}
\end{icmlauthorlist}
\icmlaffiliation{yyy}{VMware Research~}
\icmlaffiliation{xxx}{University College London}
\icmlaffiliation{zzz}{Ben-Gurion University}
\icmlaffiliation{aaa}{Stanford University}
\icmlaffiliation{bbb}{Harvard University}


\icmlkeywords{Distributed Mean Estimation, Federated Learning, Gradient Compression, Communication Efficiency.}

\vskip 0.3in
]



\printAffiliationsAndNotice{\icmlEqualContribution} 

\begin{abstract}

Distributed Mean Estimation (DME) is a central building block in federated learning, where clients send local gradients to a parameter server for averaging and updating the model. 
Due to communication constraints, clients often use lossy compression techniques to compress the gradients, resulting in estimation inaccuracies. 

DME is more challenging when clients have diverse network conditions, such as constrained communication budgets and packet losses. In such settings, DME techniques often incur a significant increase in the estimation error leading to degraded learning performance.

In this work, we propose a robust DME technique named \name that naturally handles heterogeneous communication budgets and packet losses. We derive appealing theoretical guarantees for \name and evaluate it empirically. Our results demonstrate that \name consistently improves over state-of-the-art DME techniques.

\end{abstract}

\vspace*{-2mm}
\section{Introduction}
\label{sec:introduction}

In the Distributed Mean Estimation (DME) problem, each of $n$ senders has a $d$-dimensional vector of real numbers. Each sender sends information over the network to a central receiver, who uses this information to estimate the mean of these vectors. This problem is a central building block in many federated learning scenarios, where at each training round, a parameter server averages clients' parameter updates (i.e., neural network gradients) and updates its model~\cite{mcmahan2017communication}.
As neural network gradients are often large (e.g., can exceed a billion dimensions~\cite{NIPS2012_6aca9700,shoeybi2019megatron,NEURIPS2019_093f65e0}), transmission over the network is often a bottleneck, and thus applying lossy compression to the gradients can be essential to adhere to client communication constraints, reduce the training time, and allow better inclusion and scalability.
 
Typically, the desired design property is that the receiver's resulting estimate will be unbiased. That is, the receiver's derived estimate $\hat{x}$ for a sender's vector $x \in \mathbb{R}^d$ should satisfy $\mathbb{E}[\hat{x}]=x$. Unbiasedness is attractive because, under natural conditions (including independence of estimates), as it yields a Mean Squared Error (MSE) between the mean of the received estimates and the mean of the true vectors that decays linearly with respect to the number of clients (e.g., see \citet{vargaftik2021drive}). Besides being a useful property for DME in isolation, in federated learning contexts this can remove the need for error feedback mechanisms that are commonly used to deal with biased estimates~\cite{seide20141, karimireddy2019error}, but are often not practical due to client participation patterns~\cite{kairouz2019advances}.

For unbiasedness, modern DME techniques employ randomized rounding techniques, commonly known as stochastic quantization (SQ), to map each vector coordinate to one of a limited {number of possibilities, yielding a compressed form.}

Some SQ-based techniques have known issues when used in DME. In particular, the resulting error is sensitive to the vector's distribution and the difference between the largest and smallest coordinates. This is specifically problematic in federated learning, where neural network gradients' coordinates can differ by orders of magnitude, rendering vanilla SQ inapplicable for accurate DME in many settings.

To address this limitation, recent works suggest the vector be \emph{randomly rotated} prior to stochastic quantization~\cite{pmlr-v70-suresh17a}. That is, the clients and the parameter server draw rotation matrices according to some known distribution (e.g., uniform); the clients then send the quantization of the rotated vectors while the parameter server applies the inverse rotation on the estimated rotated vector. 
Intuitively, the coordinates of a vector rotated by uniform random rotation are identically distributed (albeit weakly dependent) and are closely concentrated around their mean, leading to a small expected difference between the coordinates that allows for an accurate quantization.  For $x\in\mathbb{R}^d$, this approach achieves a Normalized $\mathit{MSE}$ ($\mathit{NMSE}$)\footnote{The normalized $\mathit{MSE}$ is the mean's estimate $\mathit{MSE}$ normalized by the mean clients' gradient squared norms (\S\ref{subsec:preliminaries}).} of $O(\frac{\log d}{n})$ using $O(1)$ bits per {coordinate per client (i.e., $O(n d)$ bits in total).}

Another approach makes use of Kashin's representation~\cite{lyubarskii2010uncertainty,caldas2018expanding,safaryan2020uncertainty}. Roughly speaking, it allows representing a $d$-dimensional vector using larger vectors with $\lambda \cdot d$ coefficients for some~$\lambda>1$, where each coefficient is smaller. Applying stochastic quantization to the Kashin coefficients allows an$\,\mathit{NMSE}$ of $O(\frac{1}{n})$ using $O(\lambda)$ bits per coordinate. Compared with~\cite{pmlr-v70-suresh17a}, using Kashin's representation yields a lower$\,\mathit{NMSE}$ at the cost of increased computational complexity~\cite{vargaftik2021drive}. 

Recent works propose algorithms that rely on clients' gradient similarity to improve guarantees. For example,~\cite{davies2021new} suggests an algorithm where if all clients' gradients have pairwise Euclidean distances of at most $a\in\mathbb R$, the resulting$\,\mathit{NMSE}$ is $O(a^2)$ using $O(1)$ bits per coordinate on average. This solution provides a good bound when gradients are similar (and thus $a$ is small). However, it may be less efficient for federated learning, where clients often have different data distributions (and thus $a$ may be large).

The recently introduced DRIVE~\cite{vargaftik2021drive} is a state-of-the-art DME algorithm that uses a single bit per coordinate. Formally, DRIVE offers an$\,\mathit{NMSE}$ of $O\parentheses{\frac{1}{n}}$ using $(1+o(1))$ bits per coordinate and improves over existing DME techniques utilizing a similar communication budget both analytically and empirically. DRIVE's improvement stems from employing a deterministic quantization instead of a stochastic one after a random rotation, yielding an asymptotic$\,\mathit{NMSE}$ improvement. DRIVE still produces unbiased estimates by adequately scaling the gradients.   

A communication budget of one bit per coordinate has been thoroughly studied~\cite{seide20141,wen2017terngrad,bernstein2018signsgd,karimireddy2019error,ben2020send,vargaftik2021drive}, and used to accelerate distributed learning systems~\cite{bytePS,HiPress}.  However, one bit per coordinate does not support many federated learning scenarios where clients have different communication budgets and network conditions.
We expand on alternative compression approaches, which are {not directly applicable to DME, in Appendix~\ref{app:extended_RW}.}

In this work, we propose \textbf{E}fficient \textbf{D}M\textbf{E} for diverse \textbf{N}etworks (\name) – a robust DME technique that supports heterogeneous communication budgets and packet loss rates. \name achieves an$\,\mathit{NMSE}$ of $O\parentheses{\frac{1}{n}}$ using $b$ bits per coordinate, for any constant $b$, including for $b < 1$, i.e., less than one bit per coordinate.  An additional feature of \name is that it naturally handles packet loss without retransmission by replacing lost coordinates with 0 values.  We extend our theoretical results to this setting for constant packet loss rates and empirically demonstrate this robustness.

\name achieves improved accuracy using a novel formalization of the quantization framework. While previous work defines the quantization via a set of quantization points, our solution requires choosing a set of \emph{intervals} whose union covers the real interval. Then,
each point is quantized to the center of mass of its interval and \emph{not} to the closest quantization point, which is counter-intuitive.
That is, our solution may quantize some points to quantization levels farther away from them than the closest. Nonetheless, such a method can reduce the entropy of the quantized vector, allowing for better $\mathit{NMSE}$ \mbox{given a communication budget.}


{We implement and evaluate \name in PyTorch~\cite{NIPS2019_9015} and TensorFlow~\cite{tensorflow2015-whitepaper}\footnotemark{} and show that \name can compress vectors with more than $67$ million} coordinates within 61~ms.
Compared with state-of-the-art DME techniques, \name consistently provides better mean estimation, which translates to higher accuracy in various federated and distributed learning tasks and scenarios.

\footnotetext{Our PyTorch and TensorFlow implementations are available as open source at \url{https://github.com/amitport/EDEN-Distributed-Mean-Estimation}.}


\section{\name}\label{sec:drive}

We start with preliminaries, overview \name, and then describe the complete details and guarantees.

\subsection{Preliminaries}\label{subsec:preliminaries}

We assume that each sender has access to randomness that is shared with the central receiver. This assumption is standard (e.g.,~\citet{pmlr-v70-suresh17a,ben2020send,vargaftik2021drive}) and can be implemented by having a shared seed for a PseudoRandom Number Generator (PRNG). Importantly, each sender uses a different seed and thus its shared \mbox{randomness is independent of that of other senders.}

Formally, we are interested in efficiently solving the DME problem. In this problem, we have a set of $n \in \mathbb N^+$ senders and a central receiver. Each sender $c\in\set{1,\ldots,n}$ has its own vector $x_c \in \mathbb R^d, \, x_c\neq 0$,\footnote{For ease of exposition,
we hereafter assume that $x_c \neq 0$ for all $c$ since this case can be handled with one additional bit. Further, in ML applications, zero gradients essentially never occur in practice.} and sends a message to the receiver. The receiver then produces an estimate of the average of these sender vectors. In particular, we focus on the setting where each sender message yields an estimate of each sender's vector $\widehat x_c$, and the receiver computes the average of the $\widehat x_c$ as an estimate of the average of the $x_c$, with the goal of minimizing its$\,\mathit{NMSE}$ defined as, 
{
$$\mathit{NMSE} \triangleq\frac{\E\brackets{\norm{\frac{1}{n}\sum_{c=1}^n \widehat x_c - \frac{1}{n}\sum_{c=1}^n x_c}_2^2}}{\frac{1}{n}\cdot\sum_{c=1}^n\norm{x_c}_2^2}~.
$$ 
}
In federated learning and other techniques based on stochastic gradient descent (SGD) and its variants (e.g., \citet{mcmahan2017communication,li2020federated,karimireddy2020scaffold}), each round includes a mean estimation of the local vectors.
Indeed, the $\mathit{NMSE}$ affects the convergence rate and often the final accuracy of the models. Further, the provable convex convergence rates for compressed SGD have a {linear dependence on the $\mathit{NMSE}$~(\citet{bubeck2015convex}, Theorem~6.3).} 

\begin{figure*}[t]
\centering
\centerline{\includegraphics[width=0.9\textwidth]{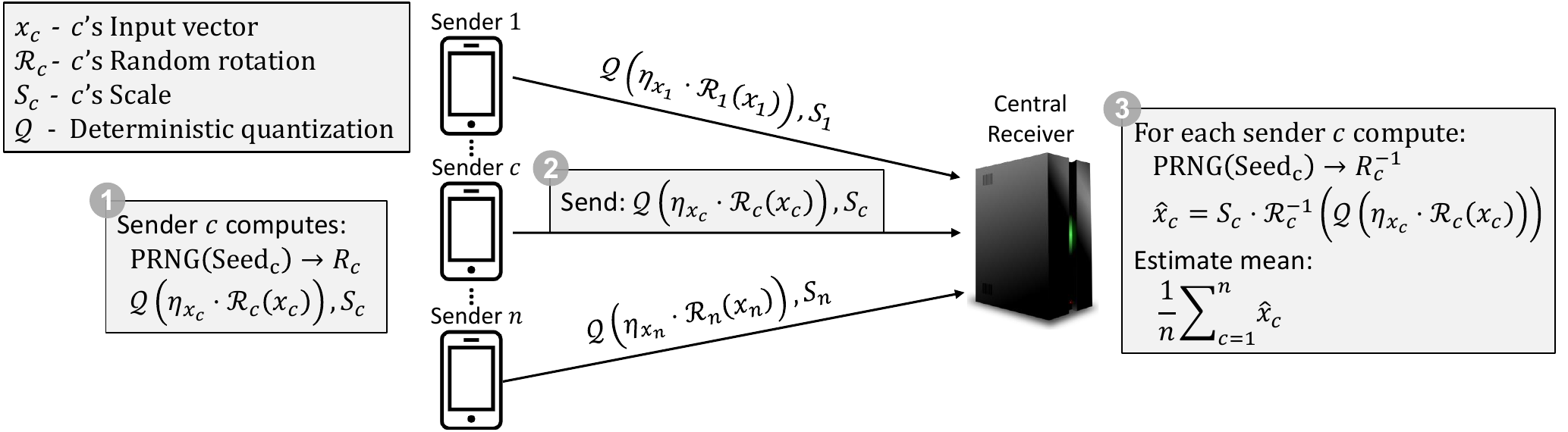}}
\caption{\name's compress and decompress methods.}
\label{fig:drive}
\end{figure*}

\subsection{\name's Overview}\label{sec:hlo_drive}

Figure~\ref{fig:drive} depicts a high-level illustration of \name.

\subsubsection{Senders.} 

To compress a vector, each sender employs three consecutive steps: rotation, quantization, and scaling.

\T{Random Rotation.} Each sender uses the shared randomness with the receiver to randomly rotate its vector and to do so \emph{independently} from other senders. Rotation can be expressed by multiplying the vector by a \emph{rotation matrix}. In particular, a rotation matrix $R\in\mathbb R^{d\times d}$ satisfies $R^TR=I$, which also implies that for any $x\in\mathbb R^d: \norm{Rx}_2=\norm{x}_2$. For ease of notation, we use $\mathcal R(x)$ to denote $Rx$ when $R$ is selected uniformly at random; in~\S\ref{sec:eval} we present an efficient implementation. Similarly, $\mathcal R^{-1}(x)$ denotes the inverse rotation, i.e., $R^{-1}x = R^Tx$. For sender $c$ and its vector $x_c \in \mathbb{R}^d$, we denote its rotated vector by $\mathcal R_c(x_c)$.

\T{Deterministic Quantization.} To encode a real-valued gradient using a finite number of bits, one must \emph{quantize} it. To design the quantization, we leverage the fact that after randomly rotating a vector, all its coordinates are identically distributed. This distribution quickly converges to a normal distribution with the vector's dimension. Specifically, for a vector $x\in\mathbb R^d$, we have that as $d$ tends to infinity, the distribution of each $\mathcal R(x)$'s coordinate tends to a normal distribution $\mathcal N(0, \frac{\norm{x}_2^2}{d})$~\cite{vargaftik2021drive}. 

Leveraging this, we can calculate the best quantization to approximate the standard normal distribution\footnotemark[\getrefnumber{footnote:shiftedBeta}] $\mathcal{N}(0,1)$ \emph{offline}.
 Then, at run time, each sender multiplies its rotated vector by a factor of $\eta_x=\frac{\sqrt{d}}{\norm{x}_2}$ and finds the best quantization for its own rotated coordinates' distribution. 
We now formalize the above, starting with defining a family of deterministic quantizations for the normal distribution.

Let $\mathcal I$ be a set of intervals with disjoint interiors such that $\cup_{I\in\mathcal I} I = \mathbb R$.
We further require two properties: 
\vspace{-0.15in}
\begin{enumerate}
    \item $\mathcal I$ is \emph{symmetric}; that is, $[a,a']\in \mathcal I {\implies} [-a',-a]\in\mathcal I$.
\vspace{-0.25in}
    \item $[-a,a]\in\mathcal{I} \implies a<1$.
\end{enumerate}

For ease of exposition, we first consider finite sets $\mathcal I$; in \S\ref{sec:entropy} and the appendix, we relax this and allow certain infinite interval families.
For example, two such partitions are $\set{(-\infty,0],[0,\infty)}$ and $\set{(-\infty,-\frac{1}{2}],[-\frac{1}{2},\frac{1}{2}],[\frac{1}{2},\infty)}$.  (Note $a,a'$ can be (minus or plus) infinity in our definition.)

Next, for each such interval $I=[a,a']\in \mathcal I$, we denote its center of mass by $q_I=\mathbb E[z | z\in I]$ where $z\sim\mathcal  N(0,1)$, i.e., $q_I=\frac{\int_a^{a'} t\cdot e^{-\frac{1}{2}t^2} dt}{\int_a^{a'} e^{-\frac{1}{2}t^2}dt}$. Also, for $z\in\mathbb R$, let $\mathcal I(z)$ denote the interval that encompasses $z$.\footnote{If $z$ is an endpoint of intervals, the one closer to zero is chosen.} 
We then define the quantization operator $\mathcal Q_{\mathcal I}(z) = q_{\mathcal I(z)}$.
When clear from context, we omit the subscript $\mathcal I$ and write $\mathcal Q$.
That is, $z$ is quantized to the center of mass of the interval in which it lies.
This definition generalizes seamlessly to vector quantization, where for $y=(y[1],\ldots,y[d])\in\mathbb R^d$ we denote 
$\mathcal Q (y) = \Big(\mathcal Q(y[1]),\ldots,\mathcal Q(y[d])\Big)$. 
Also, by the properties of $\mathcal I$, we obtain $\mathcal Q (-y)=-\mathcal Q (y)$ and $y[j]\cdot \mathcal Q (y[j]) \ge 0$ for all $j\in\set{1,\ldots,n}$ leading to $\langle y , \mathcal{Q}(y) \rangle \ge 0$ for all $y \in \mathbb{R}^d$.

For sender $c$ and its rotated vector $\mathcal R_c(x_c)$, its quantized vector is $\mathcal Q(\eta_{x_c}\cdot \mathcal R_c({x_c}))$. That is, the sender multiplies its rotated vector by $\eta_{x_c}$ before applying the quantization.

We note that it always holds that $\mathcal Q(\eta_{x_c}\cdot \mathcal R_c({x_c})) \neq 0$. Namely, the quantization process, by design, cannot nullify a client's vector. This is because $\norm{\eta_{x_c}\cdot \mathcal R_c({x_c})}_2^2=d$, which means that the absolute value of at least one coordinate is at least $1$. Thus, by the second property of $\mathcal{I}$, this coordinate cannot lie in an interval that maps it to $0$. In turn, by property 1, it also implies $\langle \mathcal R_c({x_c}) , \mathcal Q(\eta_{x_c}\cdot \mathcal R_c({x_c})) \rangle > 0$.     

In \S\ref{sec:oq}, we detail how to optimize $\mathcal I$ for different communication budgets {and how to perform the quantization efficiently.}
    
\T{Scaling.} After rotation and quantization, each sender $c$ calculates a \emph{scale} $S_c \in \mathbb{R_+}$ that is used by the receiver to scale the estimate. As we detail in \S\ref{sec:s}, scaling is the key for removing the bias introduced by the quantization.

Finally, each sender $c$ sends a representation of
$\mathcal{Q}(\eta_{x_c}\cdot\mathcal{R}_c(x_c))$ and $S_c$ to the receiver. 
This can be done with $\ceil{\log_2 |\mathcal I|} \cdot d + o(d)$ bits, i.e., using the log of the number of quantization values many bits per quantized value, and representing the scale using a sub-linear number of bits in the vector's dimension (in practice, we use a fixed number of bits, e.g., \mbox{64, to send the scale and ignore the rounding error). } 

\subsubsection{Receiver} 

The receiver reconstructs each sender's $c$ vector by first performing the inverse rotation, i.e., it uses the shared randomness to generate the same rotation matrix and computes $\mathcal R_{c}^{-1}(\mathcal Q_{c}(\eta_{x_c}\cdot\mathcal R_{c}(x_{c})))$. Then, the result is scaled by $\s_{c}$ to obtain the estimate, i.e., $\widehat x_c = S_c \cdot \mathcal{R}_c^{-1}(\mathcal{Q}(\eta_{x_c}\cdot\mathcal{R}_c(x_c)))$. Finally, the receiver averages the results from all senders and obtains the estimate of the mean, i.e., $\frac{1}{n} \sum_{c=1}^n \widehat x_c$.

\name also supports network packet losses without the need for retransmitting the lost coordinates (detailed in \S\ref{sec:variableCentroids}).

\subsection{\name's Scale}\label{sec:s}

An appealing property of \name we establish in this work is that each sender can efficiently calculate the scale $\s_{c}$ to make its estimate unbiased even though it uses a biased quantization technique. In particular, each sender $c$ uses: 
$$ S_{c}=\frac{\norm{x_{c}}_2^2}{\angles{\mathcal R_{c}(x_{c}), \mathcal Q(\eta_{x_c}\cdot\mathcal R_{c}(x_{c}))}}~. $$
With this scale, we obtain the following formal guarantee whose proof appears in Appendix~\ref{app:unbiased}.

\begin{restatable}{theorem}{exentedDriveUnbiasedness}\label{thm:exentedDriveUnbiasedness}
For all  $x \in \mathbb{R}^d$, using \name with the scale
{$\s=\frac{\norm{x}_2^2}{\angles{\mathcal R(x), \mathcal Q(\eta_{x}\cdot\mathcal R(x))}}$ 
results in $\E [\widehat x] = x$~.}
\end{restatable}
Intuitively, this scale ensures that the reconstructed vector $\widehat x_c$ lies on a hyperplane tangent to the original vector's $x_c$ point on the sphere. 
Since the rotation has no preferred direction, the expected value of the reconstructed vector produces precisely the original one. Specifically, the proof relies on this property by showing that for each rotation, there exists a matching rotation with the same bias but the opposite sign, and each such pair's average yields the original vector. 

\subsection{\name's$\,\mathit{NMSE}$}\label{sec:NMSE}

We start with the following definition. For estimation of a single vector $x$, we define the vector-$\mathit{NMSE}$ ($\mathit{vNMSE}$) as

$$\mathit{vNMSE} \triangleq \frac{\mathbb E\brackets{\norm{x-\widehat x}_2^2}}{\norm{x}_2^2}~.$$ 

For a sender $c$, we use $\mathit{vNMSE}(c)$ to denote its $\mathit{vNMSE}$. Now, since the estimates of the sender vectors are independent (as senders sample their rotation matrices independently) and unbiased (according to Theorem~\ref{thm:exentedDriveUnbiasedness}), we obtain the following result whose proof appears in Appendix~\ref{app:vnmsetonmselemma}.

\begin{restatable}{lemma}{vnmsetonmselemma}\label{thm:vNMSE_to_NMSE}
Consider $n$ senders. It holds that
$$
\mathit{NMSE} = \frac{\sum_{c=1}^n \mathit{vNMSE}(c) \cdot \norm{x_c}_2^2}{n \cdot \sum_{c=1}^n \norm{x_c}_2^2}~.
$$
\end{restatable}

Observe that for the special case where all senders use the same set $\mathcal I$, it holds that $\mathit{NMSE} = \frac{1}{n} \cdot \mathit{vNMSE}$ since $\mathit{vNMSE}(c)$ is the same for all senders.

\looseness=-1
Accordingly, we obtain a bound on the $\mathit{NMSE}$ by bounding each client's $\mathit{vNMSE}$ using the following theorem, whose proof appears in Appendix~\ref{app:nmse}.
The proof relies on a novel mathematical framework that leverages the fact that the rotated vector's distribution is that of a vector of independent $\mathcal N(0,1)$ random variables $Z\in\mathbb R^d$ multiplied by the input vector's norm and divided by $Z$'s norm~\cite{vargaftik2021drive}. 
Specifically, we define events that control quantities of interest (e.g., that the norm of $Z$ is highly concentrated around its mean). We then show that these events hold with high probability and infer that our $\mathit{vNMSE}$ converges to the following function of the \mbox{quantization error of a single $\mathcal N(0,1)$ variable.}

\begin{restatable}{theorem}{vnmsetheorem}\label{thm:nmse}
Let $z \sim \mathcal{N}(0,1)$. For all $x \in \mathbb{R}^d$, with
$\s=\frac{\norm{x}_2^2}{\angles{\mathcal R(x), \mathcal Q(\eta_{x}\cdot\mathcal R(x))}}$, \name~satisfies:
\begin{align*}
&\mathit{vNMSE} \le \frac{1}{\mathbb{E}\Big[\para{{\mathcal{Q}}(z)}^2\Big]} - 1 + O\parentheses{\sqrt{\frac{\log d}{d}}}.
\end{align*}
\end{restatable}

Also, $\E\sbrac{z\mathcal{Q}(z)} {=}~ \E\sbrac{\mathcal{Q}(z)\cdot\sbrac{\E\sbrac{z} \big| \mathcal{Q}(z)}} {=}~ \E\sbrac{\para{\mathcal{Q}(z)}^2}$ and thus $\mathbb{E}\Big[\para{{\mathcal{Q}}(z)}^2\Big] {=}~ 1 ~{-}~ \mathbbm{E}\big[(z{-}\mathcal{Q}(z))^2\big]$.
This means that minimizing the $\mathit{vNMSE}$ bound is achieved by minimizing the quantization's $\mathit{MSE}$ with respect to $z \sim \mathcal{N}(0,1)$.

Next, we obtain the following corollary.
\begin{corollary}\label{corr:nmse}
For $d\to\infty$, the $\mathit{vNMSE}$ upper bound in Theorem \ref{thm:nmse} approaches  
$\frac{1}{\mathbb{E}\sbrac{\para{{\mathcal{Q}}(z)}^2}} - 1$, where $z \sim \mathcal{N}(0,1)$.
\end{corollary}
Our proofs have focused on the upper bound, but Corollary~\ref{corr:nmse} is tight;  our proofs could be extended to a lower bound, and our experiments coincide with this claim.

We later give examples of these guarantees and how they relate to the number of bits used per coordinate in \S\ref{sec:oq}.

\section{Optimal $2^b$-Values Quantization}\label{sec:oq}


With $b$ bits per coordinate, we can use $2^b$ quantization values. Since our goal is to minimize the $\mathit{MSE}$ from a standard normal random variable to its quantization, we \emph{precalculate} the optimal quantization for $2^b$ values using the known \emph{Lloyd-Max Scalar Quantizer}~\cite{lloyd1982least,max1960quantizing} for the normal distribution.\footnote{One can slightly lower the quantization error by optimizing the quantization values for the actual distribution of the rotated coordinates (which is a shifted Beta distribution~\cite{vargaftik2021drive}). However, as mentioned there, this distribution approaches the normal distribution rapidly as $d$ grows (e.g., the difference is negligible even for $d$ of several hundred), and our focus is on federated learning where $d$ is considerably larger (e.g., millions).\label{footnote:shiftedBeta}}


For ease of exposition, for an integer bit budget $b\in\mathbb N^+$, we denote by $\mathcal I_b$ the optimal set of $2^b$ intervals and by $Q_{\mathcal I_b}$ the resulting quantization values. 
For example, the intervals and quantization values for $b=1$ and $b=2$ are: 
\resizebox{\columnwidth}{!}{
\begin{minipage}{\linewidth}
\begin{align*}
    & \mathcal I_1 = \set{(-\infty,0],[0,\infty)}\ , \quad Q_{\mathcal I_1} = \set{\pm \sqrt{\frac{2}{\pi}}}\approx\{\pm 0.79788\}. \\ 
    & \mathcal I_2 \approx \{(-\infty,-0.9816],[-0.9816,0],[0,0.9816], [0.9816,\infty)\},\\ 
    &  Q_{\mathcal I_2} \approx \{\pm 0.45278, \pm 1.51042\}.
\end{align*}
\end{minipage}
}

We note that when $Q$ is built using the Lloyd-Max quantizer, the quantization can be efficiently computed by 
$$
\mathcal Q(x) = \text{argmin}_{y\in  Q^d}\norm{ \sqrt{d}\cdot \frac{x}{\norm{x}_2} - y}_2~.
$$

Namely, for such quantization value, the center of mass of a scaled coordinate's interval is also the closest quantization value to that coordinate. 
For clarity, we now show how Corollary \ref{corr:nmse} applies for $Q_{\mathcal I_1}$ and $Q_{\mathcal I_2}$.

\textbf{Example 1.} For $Q_{\mathcal I_1}$ we obtain
\begin{align*}  
&\frac{1}{\mathbb{E}\Big[ \para{{\mathcal{Q}_{\mathcal{I}_1}}(z)}^2\Big]} - 1 = \frac{1}{\sum_{I\in\mathcal I_1} q_I^2 \cdot \mathbb{P}(z \in I)} - 1 = \\
&\frac{1}{\frac{1}{2}\para{\sqrt{\frac{2}{\pi}}}^2 + \frac{1}{2}\para{-\sqrt{\frac{2}{\pi}}}^2} - 1 =\frac{1}{\frac{2}{\pi}}-1\approx 0.571.
\end{align*}

That is, as $d \rightarrow \infty$, the $\mathit{vNMSE}$ goes to approximately $0.571$, which coincides with the corresponding result for DRIVE~\cite{vargaftik2021drive}. In fact, without coordinate losses (\S\ref{subsec:losses}), using \name with $Q_{\mathcal I_1}$ is equivalent to using DRIVE since $S\cdot\mathcal{Q}_{\mathcal{I}_1}(\eta_{x}\cdot\mathcal{R}(x))=\frac{\norm{x}_2^2}{\norm{\mathcal{R}(x)}_1}\cdot\operatorname{sign}(\mathcal{R}(x))$~. 

\textbf{Example 2.} For $Q_{\mathcal I_2}$ we have that
\begin{multline*}
\hspace*{-2mm}\Pr\sbrac{z\in \sbrac{0,0.9816}} 
= \frac{1}{\sqrt{2\pi}}\int_0^{0.9816} e^{-t^2/2}dt \approx 0.33685.
\end{multline*}
Therefore, we obtain: 
\begin{align*}
&\frac{1}{\mathbb{E}\Big[ \para{{\mathcal{Q}_{\mathcal{I}_2}}(z)}^2\Big]} - 1 = \frac{1}{\sum_{I\in\mathcal I_2} q_I^2 \cdot \mathbb{P}(z \in I)} - 1 \approx \\
&\frac{1}{2 \cdot 0.33685 \cdot (0.45278)^2 + 2 \cdot 0.16315 \cdot (1.51042)^2} - 1\\
& \approx \frac{1}{0.88228} - 1 \approx 0.134,
\end{align*}
which is an improvement of by more than a factor of $4$ in comparison to Example 1. 

In practice, we find that the empirical $\mathit{vNMSE}$ (and the resulting$\,\mathit{NMSE}$) match that of Corollary \ref{corr:nmse} in all our experiments, for any $d$ that is larger than a few hundreds.  

\section{Handling Heterogenity and Loss}\label{sec:variableCentroids}

We next detail how \name operates with general bit budget constraints and lossy networks, and discuss its compatibility with variable-length encoding techniques.

\subsection{Heterogeneous Sender Bit Budget} \label{sec:heteroBits}

Often in federated learning, senders may have different resource constraints, particularly networking constraints~\cite{nishio2019client}. Therefore, it is beneficial for an algorithm to allow senders to use different amounts of compression, tuned to their own available communication budget. Accordingly, we provide two generalizations that maintain the strong guarantees of Theorems \ref{thm:exentedDriveUnbiasedness} and \ref{thm:nmse} and allow \name's senders to adapt their bit budget per coordinate.  Specifically, we allow each sender to use its own set of quantization values and to use a non-integer number of bits $b$ per coordinate (in expectation).

\textbf{Super-bit compression ($b\ge 1$). } For a sender who wishes to use an integer $b$ bits per coordinate, we simply use $Q_{\mathcal I_b}$.  For non-integer $b > 1$, we propose the following generalization. We quantize each coordinate using $Q_{\mathcal I_{\floor{b}+1}}$ with probability $b-\floor{b}$, and with $Q_{\mathcal I_{\floor{b}}}$ with probability $1 - (b-\floor{b})$. This means that each client's quantization is a distribution over $Q_{\mathcal I_{\floor{b}+1}}$ and $Q_{\mathcal I_{\floor{b}}}$. 
The choice of which coordinates to send using more bits are selected using shared randomness (to simulate independent weighted coin flips). In practice, this means the actual bit usage may slightly deviate from (but is concentrated around) its expected value of $b$. This approach avoids introducing additional overhead from needing to communicate this information explicitly (although $b$ does need to be sent or otherwise agreed upon).  We observe that Theorems \ref{thm:exentedDriveUnbiasedness} and \ref{thm:nmse} hold for this scenario (the proofs are in Appendix \ref{app:unbiased} and \ref{app:nmse} respectively). 

For example, for $b=1.5$, each coordinate is quantized using $Q_{\mathcal I_1}$ or $Q_{\mathcal I_2}$ with equal probability. The resulting $\mathit{vNMSE}$ for this case is (for $d\to\infty$):
\begin{align*}  
& \frac{1}{\frac{1}{2} \cdot  \mathbb{E}\big[ \para{{\mathcal{Q}}_{\mathcal I_1}(z)}^2\big] + \frac{1}{2} \cdot  \mathbb{E}\big[ \para{{\mathcal{Q}}_{\mathcal I_2}(z)}^2\big]}-1
 \\
&\approx \frac{1}{\frac{1}{2}\cdot (\frac{2}{\pi})  + \frac{1}{2}\cdot (0.88228)} - 1 \approx 0.317.    
\end{align*}

We note that the naive solution of simply dividing the vector into two halves and sending each separately (one half using \name with $Q_{\mathcal I_1}$ and the other with $Q_{\mathcal I_2}$) yields a higher $\mathit{vNMSE}$ for the reconstructed vector (i.e., $\frac{1}{2}\cdot 0.571+ \frac{1}{2}\cdot 0.134 = 0.352$ instead of $0.317$).


\textbf{Sub-bit compression ($b < 1$). } For sub-bit compression, we use $Q_{\mathcal I_1}$  with random sparsification.
%
%
Formally, the random sparsification procedure has only a single parameter $p\in(0,1]$. For a vector $x \in \mathbb{R}^d$, its random sparsification is $\frac{1}{p} \cdot m_{rs} \circ x$, where $m_{rs}\in\{0,1\}^d$ is a uniformly random sample of size $\norm{m_{rs}}_1=d\cdot p$ and $\circ$ stands for the coordinate-wise product (i.e., $m_{rs}$ is a random mask). For random sparsification, it holds that
$$\mathit{vNMSE} = \frac{{\sum_{i=1}^d (1-p)x[i]^2+p(\frac{1}{p}-1)^2x[i]^2}}{\norm{x}_2^2}=\frac{1}{p}-1~.
\vspace*{-1mm}
$$

There are two ways to apply the random sparsification: before or after the rotation. In practice, we find that the resulting $\mathit{vNMSE}$ is similar in both cases and use the sparsification prior to rotation which is more efficient since it reduces the dimension of the compressed vector. 

For example, for $b=0.7$, we sparsify uniformly at random $30$\% of the coordinates (i.e., set $p=0.7$), multiply the remaining coordinates by a factor of $\frac{1}{0.7}$ to preserve unbiasedness, and then compress them using \name with $Q_{\mathcal I_1}$ (i.e., a single bit per coordinate). In turn, the receiver decodes the compressed sparsified vector and then restores the original using the same random mask (i.e., generating the same random mask using the shared randomness). 

We strengthen the above choice using the following formal result whose proof appears in Appendix~\ref{app:sub-bit}.

\begin{restatable}{lemma}{twocompressorsvNMSE}\label{lemma:twocompressorsvNMSE}
Consider two unbiased compression techniques $\mathcal{A}$ and $\mathcal{B}$ (i.e., $\forall x:\mathbb E[\mathcal A(x)]=\mathbb E[\mathcal B(x)]=x$) with independent randomness. Then,
\begin{enumerate}
    \item $\forall x:\frac{\E\sbrac{\norm{x-\mathcal{A}(x)}}_2^2}{\norm{x}_2^2} \le A \text{~~and~~} \frac{\E\sbrac{\norm{x-\mathcal{B}(x)}}_2^2}{\norm{x}_2^2} \le B \implies \forall x:\frac{\E\sbrac{\norm{x-\mathcal{B}(\mathcal{A}(x))}}_2^2}{\norm{x}_2^2} \le A + AB + B~.$
    \item $\forall x:\frac{\E\sbrac{\norm{x-\mathcal{A}(x)}}_2^2}{\norm{x}_2^2} \ge A \text{~~and~~} \frac{\E\sbrac{\norm{x-\mathcal{B}(x)}}_2^2}{\norm{x}_2^2} \ge B \implies \forall x:\frac{\E\sbrac{\norm{x-\mathcal{B}(\mathcal{A}(x))}}_2^2}{\norm{x}_2^2} \ge A + AB + B~.$
\end{enumerate}

\end{restatable}

Accordingly, we get that \name with sparsification has $\mathit{vNMSE}\le \frac{\pi}{2\cdot p}-1 + O\parentheses{\sqrt{\frac{\log d}{d\cdot p^2}}}$ and obtain the following.

\begin{corollary}\label{corr:subitsim}
\name's $\mathit{vNMSE}$ with constant $b\in(0,1]$ bits per coordinate satisfies
\begin{align*}
&\lim_{d\to\infty} \mathit{vNMSE} = \frac{\pi}{2\cdot b}-1~.
\end{align*}
\end{corollary}

For example, using $b=0.1$ yields a $\mathit{vNMSE}$ of $\approx14.707$. More generally, for any fixed bit budget $b>0$ we have that $\mathit{vNMSE} = O(1)$ and thus we get $\mathit{NMSE}= O(\frac{1}{n})$~.

\subsection{Lossy Networks}\label{subsec:losses}

Distributed and federated learning systems (e.g.,~\citet{HiPress,bytePS}) typically assume reliable packet delivery, e.g., using TCP to retransmit lost packets or using RDMA or RoCEv2, which rely on a lossless fabric. However, it is useful to design algorithms that can cope with packet loss on standard IP networks. Indeed, a recent effort by~\citet{ye2021decentralized} extends SGD to support packet loss.

We model packet loss as a sparsification with parameter $p\in(0,1]$, so a fraction $p$ of the coordinates arrive at the receiver. Hence this modeling is similar to that of the sub-bit regime, but there are inherent differences:
(1) \emph{The sparsification may not be random}. Instead, we assume an oblivious adversary may pick any subset of packets to drop. That is, the adversary may choose to drop packets based only on the packet indices, without any knowledge about the content of the packets;
(2) The sparsification is done after the rotation;
(3) The quantization scheme is not restricted to using ${\mathcal I_1}$.  

To support this scheme, no changes at the sender are required. For the receiver, before employing the inverse rotation and scaling, it simply treats any lost coordinates as $0$ and multiplies the reconstructed rotated vector by $\frac{1}{p}$ (the receiver determines $p$ by counting the number of received coordinates). This scheme preserves our theoretical guarantees due to \mbox{the following Lemma, which is proven in Appendix~\ref{app:lossy}.}

\begin{restatable}{lemma}{lossylemma}\label{lemma:lossylemma}
Let $x\in\mathbb{R}^d$ and let $m_{ds}\in\{0,1\}^d$ be a \emph{deterministic} mask. 
Denote $p = \frac{\norm{m_{ds}}_1}{d}$ and let $\mathcal{R}_{ds}(x) = \frac{1}{p} \cdot m_{ds} \circ \mathcal{R}(x)$.  Then, using \name with $\mathcal{R}_{ds}(x)$ instead of $\mathcal{R}(x)$ results in:
\begin{enumerate}[topsep=0pt,itemsep=-1ex,partopsep=1ex,parsep=1ex]
    \item $\E\sbrac{\widehat x} = x$~.
    \item $\mathit{vNMSE} \le \frac{1}{p \cdot \mathbb{E}\sbrac{\para{{\mathcal{Q}}(z)}^2}} - 1 + O\parentheses{\sqrt{\frac{\log d}{d\cdot p^2}}}$~.
\end{enumerate}
\end{restatable}

Lemma~\ref{lemma:lossylemma} shows that when performing DME, we can settle for partial information from all the senders and can replace missing information by 0s and rescale each vector accordingly.  Of course, this increases the error, but Lemma~\ref{lemma:lossylemma} bounds this increase. This feature enables the use of lossy transport protocols such as UDP (e.g., as proposed by~\citet{ye2021decentralized}), or allows the receiver to avoid waiting for retransmissions of lost packets.  Hence we can trade-off error and delay (latency) by allowing partial information.  Another benefit is that lossy protocols often have reduced overheads (e.g., smaller headers) since they do not require maintaining state and reliable delivery.

\subsection{Variable-Length Encoding}\label{sec:entropy}

A standard approach to compressing vectors with a small number of possible values that are unequally distributed is using entropy encoding methods such as Huffman~\cite{huffman1952method} or arithmetic encoding~\cite{pasco1976source,rissanen1976generalized}.
Intuitively, the number of times each quantization value appears in $\mathcal Q(\eta_x \cdot \mathcal R(x))$ may not be $\frac{d}{|\mathcal I|}$.
Indeed, for $\mathcal I_b$ with $b\ge 2$, {the probability of the values is not equal.}

Formally, denoting by $p_I=\mathbbm{P}(z\in I)$ the probability that a normal variable $z\in\mathcal N(0,1)$ lies in the interval $I$, $H_{\mathcal I}\triangleq\sum_{I\in\mathcal I} -p_I\log_2(p_I)$ is the \emph{entropy} of the distribution induced by $\mathcal I$.
For example, as discussed in Example~2 (\S\ref{sec:oq}), for $\mathcal{I}_2$, the intervals $[-0.9816,0],[0,0.9816]$ have probability that is more than double than that of $(-\infty,0.9816], [0.9816,\infty)$. The {entropy}  of the distribution is
$H_{\mathcal I_2}\approx 1.91$ bits.  
This suggests that, for a large enough dimension $d$, we should be able to compress the vector to at most $H_{\mathcal I_2}+\epsilon$ bits per coordinate for any constant $\epsilon$.

Such an encoding may also allow us to use more quantization values for the same bit budget. For example, using the Lloyd-Max Scalar Quantizer~\cite{lloyd1982least,max1960quantizing} with $9$ quantization values, we get an entropy of $\approx 2.98$ bits, which would allow us to use three bits per coordinate for large enough vectors. In particular, this would reduce the $\mathit{vNMSE}$ by nearly 20\%, compared to $\mathcal I_3$, at the cost of additional computation.

Taking this a step further, it is possible to consider the resulting entropy when choosing the set of intervals.
To optimize the quantization, we are looking for a set $\mathcal I$ that maximizes ${\E\brackets{{\para{\mathcal Q_{\mathcal I}(z)}}^2}}$  such that its entropy is bounded by $b$.
This problem is called Entropy-Constrained Vector Quantization (ECVQ)~\cite{chou1989entropy}.
The algorithm proposed in~\citet{chou1989entropy} has several tunable parameters that may affect the output of the algorithm.
We implemented the algorithm and scanned a large variety of parameter values.
For $b=3$, for example, the best obtained $\mathit{vNMSE}$ is $\approx 0.02274$, compared
to an $\mathit{vNMSE}$ of $\approx0.03572$ for \name with $\mathcal I_3$ and without entropy encoding.

We propose a simpler approach that is more computationally efficient.
Given a bandwidth constraint $b>0$, let \mbox{$\Delta_b>0$} denote the smallest real number such that $H_{\mathcal I_{\Delta_b}}\le b$ where $\mathcal I_{\Delta_b}=\set{[\Delta_b\cdot (n-\frac{1}{2}),\Delta_b\cdot (n+\frac{1}{2})] \mid n\in\mathbb Z}$.{This choice of $\mathcal{I}$ respects the properties that are required by Theorems \ref{thm:exentedDriveUnbiasedness} and \ref{thm:nmse} for any fixed $\Delta_b \in (0,2)$.}

For example, using $b=3$, we have $\Delta_3 \approx 0.5224$. Then, the resulting $\mathit{vNMSE}$ is $\approx0.022741$ (within $0.1$\% from the ECVQ solution), an improvement of $\approx20\%$ over the Lloyd-Max Scalar Quantizer with $9$ quantization values (which we can encode with $b=3$ bits per coordinate by applying entropy encoding) and of $\approx36\%$ over $8$ quantization values (which is encodable with $b=3$ without entropy encoding). 
A benefit of our approach is that it allows computing the quantization much faster as each $a\in\mathbb R$ is efficiently mapped into the interval $\brackets{\Delta_b \cdot \para{\Big\lfloor{\frac{a}{\Delta_b}\Big\rceil}-\frac{1}{2}} , \Delta_b \cdot \para{\Big\lfloor{\frac{a}{\Delta_b}\Big\rceil}+\frac{1}{2}}}$, where $\lfloor\cdot\rceil$ rounds to the nearest integer (i.e., the interval's index is $n=\Big\lfloor{\frac{a}{\Delta_b}\Big\rceil}$).

Rate-distortion theory implies a lower bound on $\mathbbm{E}\big[(z{-}\mathcal{Q_{\mathcal I}}(z))^2\big]$ for any quantization interval set $\mathcal I$~\cite{cover1999elements}. Specifically, it implies that $\mathbbm{E}\big[(z{-}\mathcal{Q_{\mathcal I}}(z))^2\big]\ge 4^{-b}$ for any $\mathcal I$ such that $H_{\mathcal I}\le b$. 
Note that since $\mathbbm{E}\big[(\mathcal{Q_{\mathcal I}}(z))^2\big] = 1-\mathbbm{E}\big[(z{-}\mathcal{Q_{\mathcal I}}(z))^2\big]$, we get that the lower bound on the $\mathit{vNMSE}$ attainable using any quantization is $\frac{1}{1-4^{-b}}-1=\frac{4^{-b}}{1-4^{-b}}$.
Also, in Appendix~\ref{app:entropyEval}, we illustrate the different quantization values and their expected squared error (Figure~\ref{fig:EDEN_vNMSE_EE}) and show that our $\mathcal I_{\Delta_b}$ approach is close to the lower bound while allowing computationally efficient quantization.
In practice, the entropy $H_{\mathcal I}$ may deviate from its expectation since the frequency of a quantization value, $q_I$, may not be exactly $p_I\cdot d$.
However, due to concentration, this has minimal effect.  
We elaborate on this in Appendix~\ref{app:entropy}.

\iffalse
\begin{figure}[t]
     \centering
     \subfloat[$b=2$ bit budget.]{
         \includegraphics[width=\linewidth]{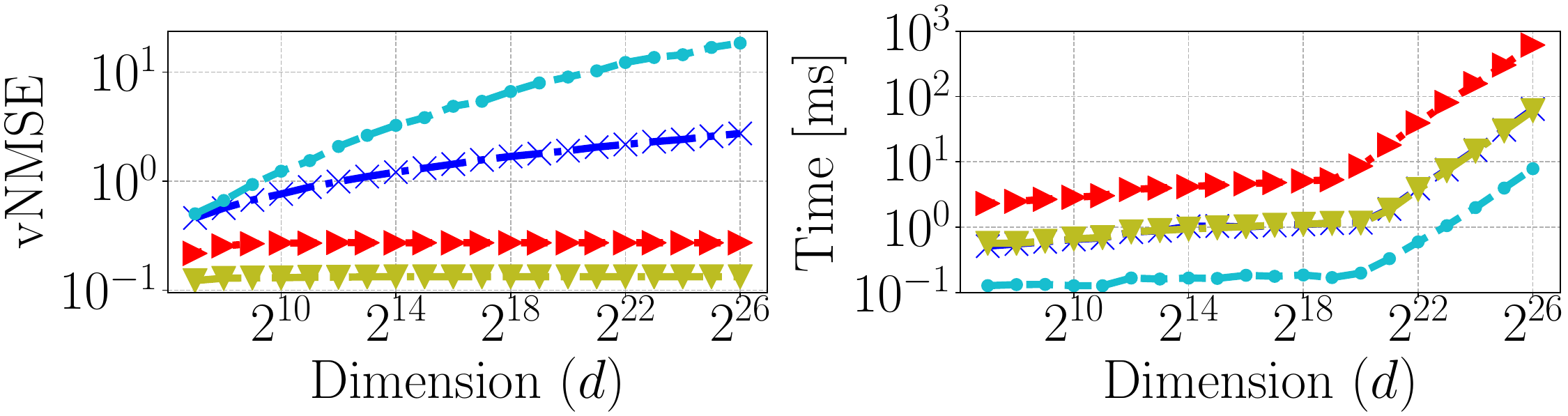}
         \label{fig:nmse:normal2}
     }\\\vspace{-1mm}
     \subfloat[$b=4$ bit budget.]{
         \includegraphics[width=\linewidth]{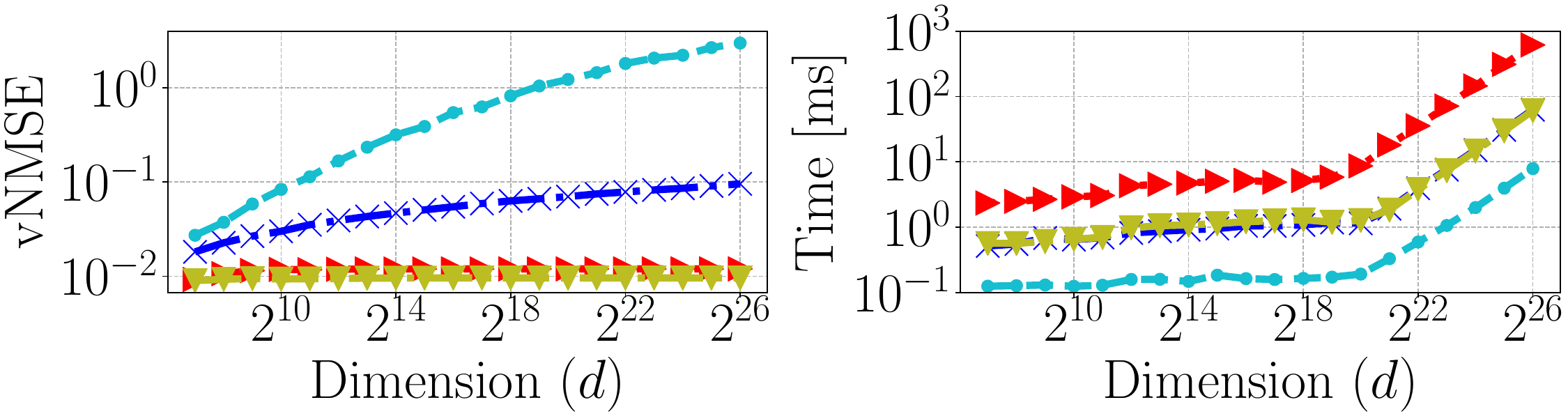}
         \label{fig:nmse:normal4}
     }\\ \vspace{-1mm}
     \subfloat[$b=6$ bit budget.]{
         \includegraphics[width=\linewidth]{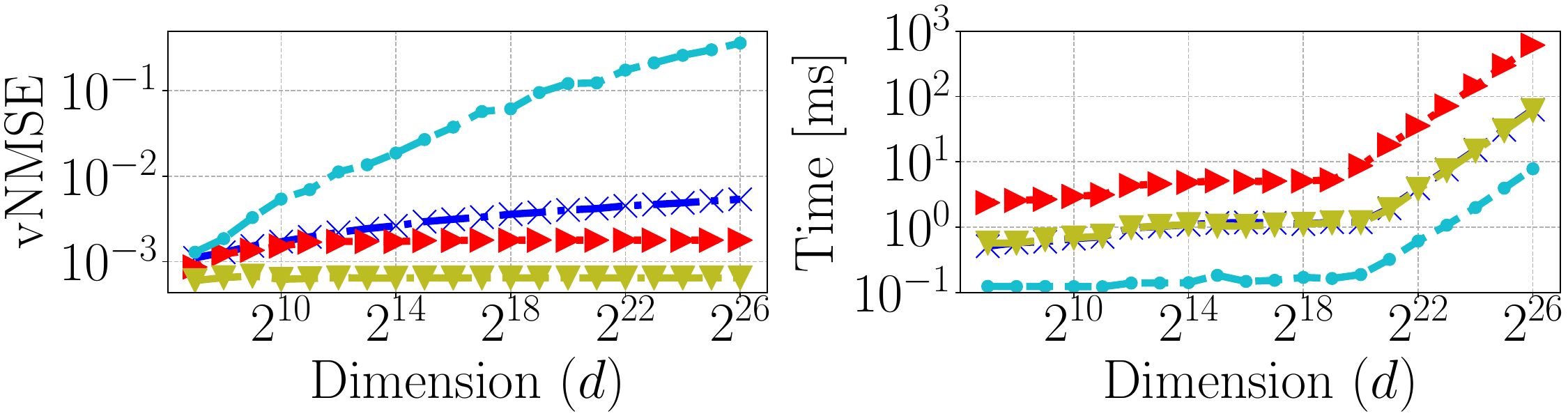}
         \label{fig:nmse:normal6}
     } \\
     \vspace{-1mm}
     \subfloat{
         \centering
         \includegraphics[width=.7\linewidth]{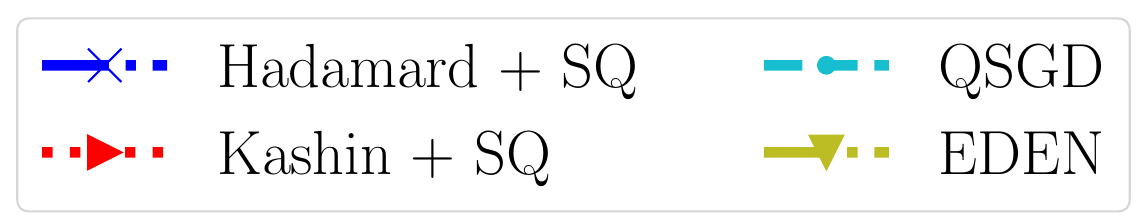}
         \label{fig:nmse:normalLegend2c}
     }
     \vspace{-1mm}
     \caption{The $\mathit{vNMSE}$ and compression time as a function of the dimension $d$ for LogNormal(0,1) distribution.}
     \label{fig:fast_and_accurate_main}
\end{figure}
\else
\begin{figure*}[t]
     \centering
     \subfloat[$b=2$ bit budget.]{
         \includegraphics[width=.323\linewidth]{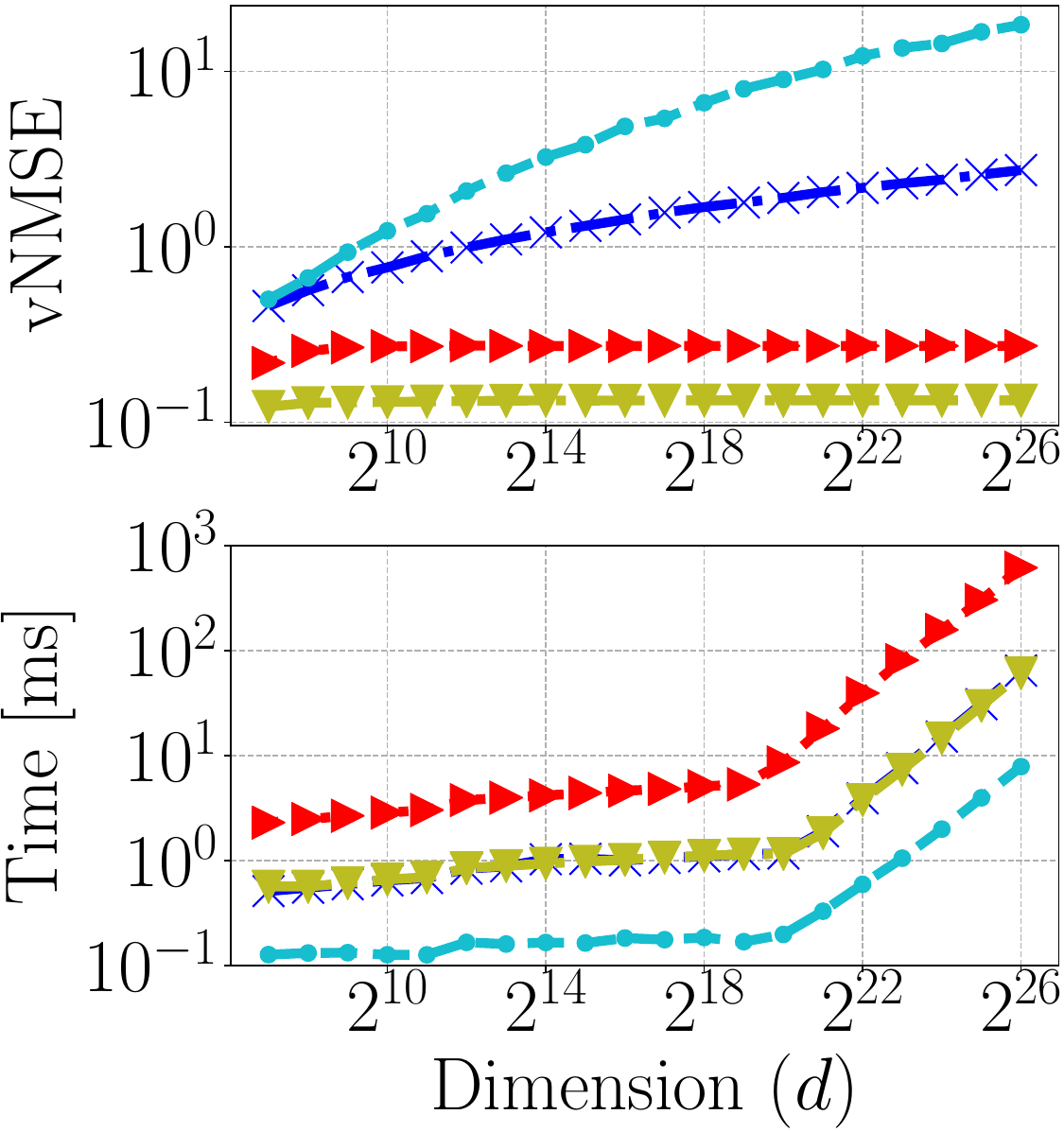}
         \label{fig:nmse:normal2}
     }
     \subfloat[$b=4$ bit budget.]{
         \includegraphics[width=.323\linewidth]{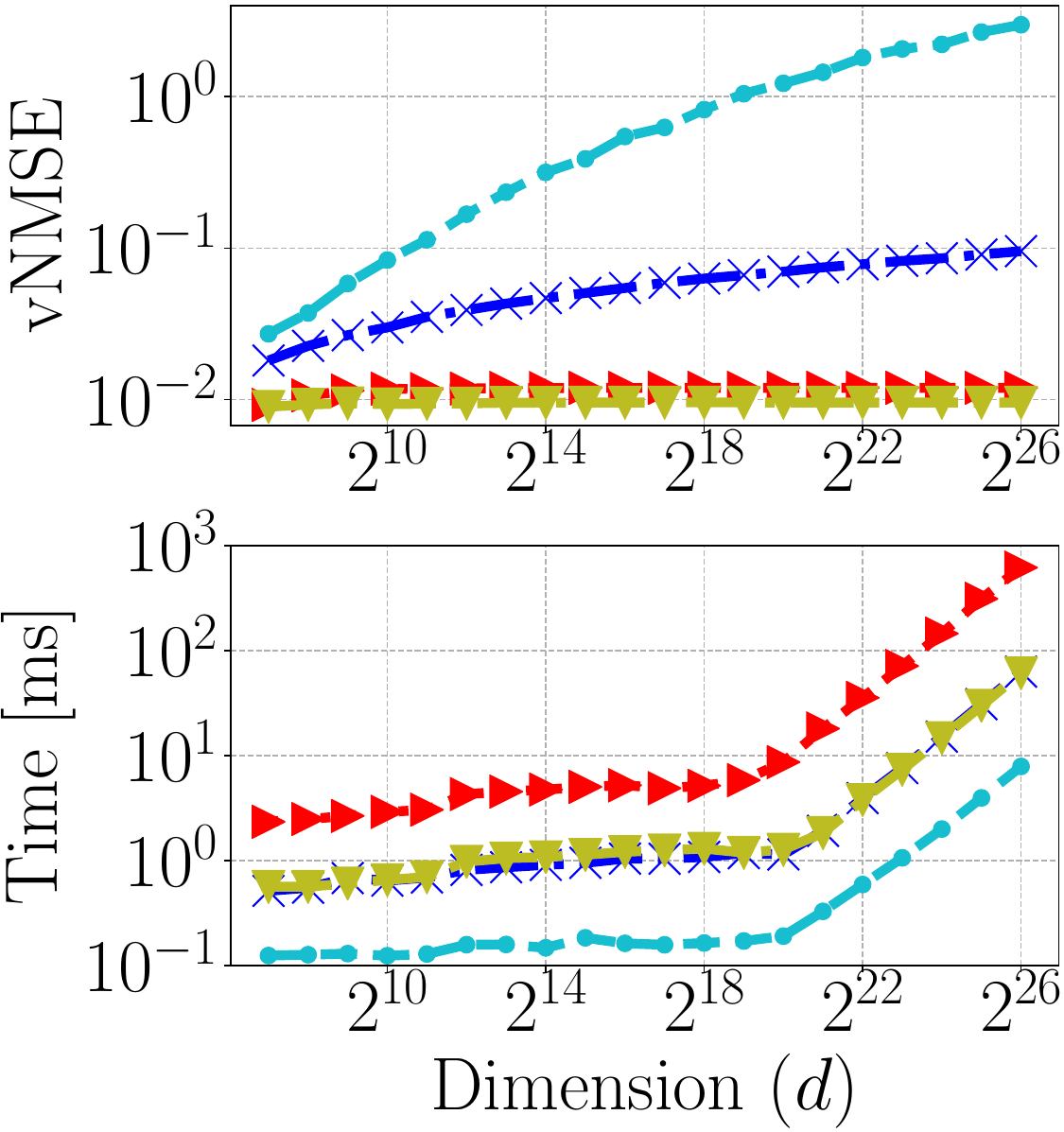}
         \label{fig:nmse:normal4}
     }
     \subfloat[$b=6$ bit budget.]{
         \includegraphics[width=.323\linewidth]{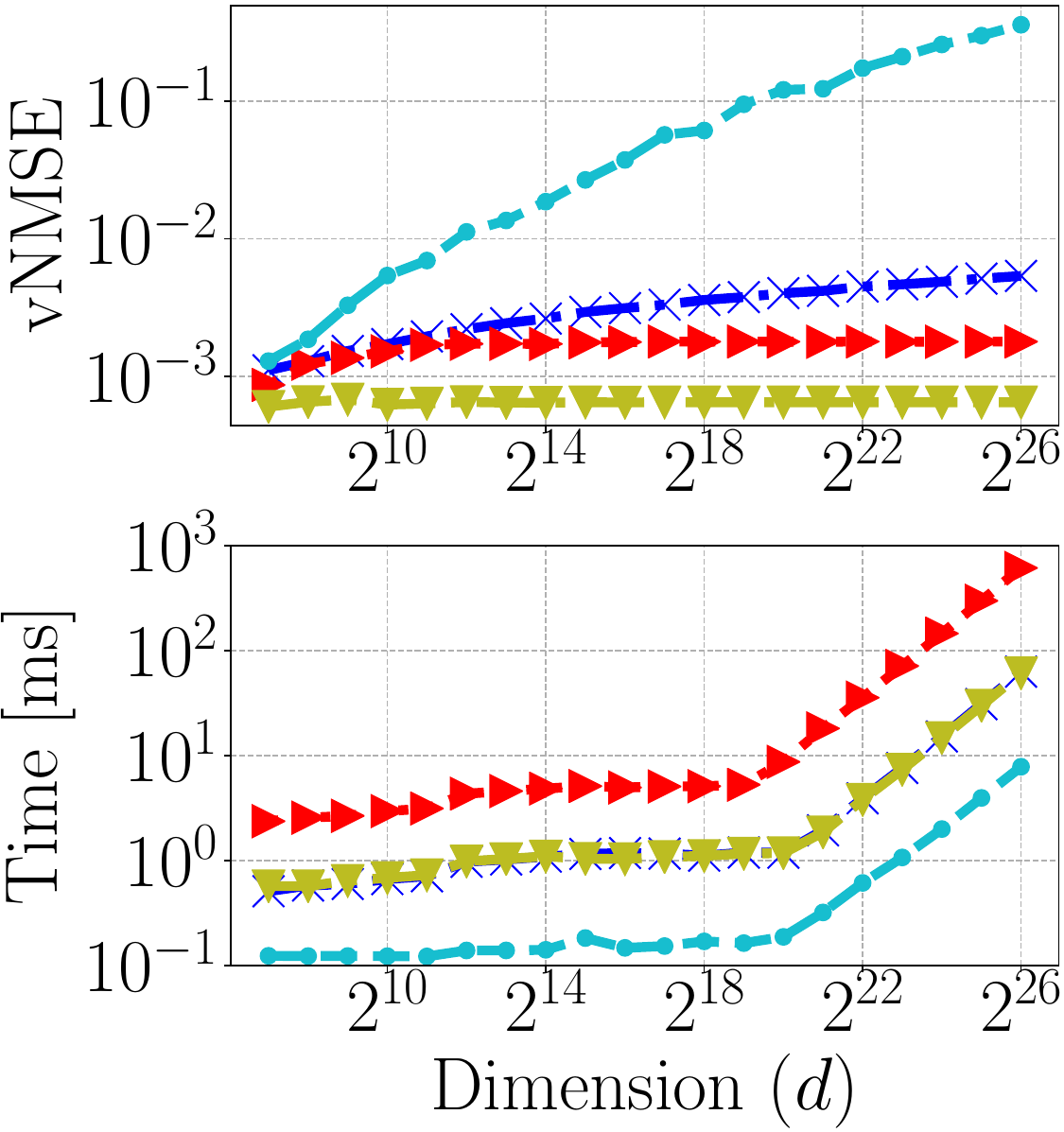}
         \label{fig:nmse:normal6}
     } \\
     \vspace{-2mm}
     \subfloat{
         \centering
         \includegraphics[width=.7\linewidth]{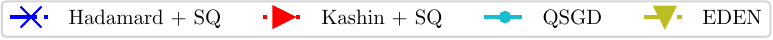}
         \label{fig:nmse:normalLegend2c}
     }
     \vspace{-3mm}
     \caption{The $\mathit{vNMSE}$ and compression time as a function of the dimension $d$ for LogNormal(0,1) distribution.}
     \label{fig:fast_and_accurate_main}
     \vspace{-3mm}
\end{figure*}
\fi
\vspace*{2mm}
\section{Evaluation}\label{sec:eval}
\vspace*{2mm}
We evaluate \name using different federated and distributed learning tasks.
We compare with a non-compressed baseline that uses 32-bit floating-point representation for each coordinate (Float32) and the following DME techniques: (1) Stochastic quantization (SQ) applied after the randomized Hadamard transform (Hadamard + SQ)~\cite{pmlr-v70-suresh17a,konevcny2018randomized}\footnote{SQ~\cite{barnes1951electronic, doi:10.1137/20M1334796} normalizes the vector into the range $[0, 2^b-1]$ (using min-max normalization), adds uniform noise in $(-0.5, 0.5)$, and then rounds to the nearest integer. thus providing an unbiased estimate of each coordinate.}; (2) SQ applied over the vector's Kashin's representation (Kashin + SQ)~~\cite{lyubarskii2010uncertainty,caldas2018expanding,safaryan2020uncertainty}; and (3) QSGD \cite{NIPS2017_6c340f25}, which normalizes the input vector by its euclidean norm and separately sends its sign and its quantized (using SQ) absolute values. 

We exclude methods that involve client-side memory since these can often work in conjunction with all tested methods (e.g.,~\citet{karimireddy2019error, richtarik2021ef21}) and are less applicable in cross-device federated scenarios~\cite{kairouz2019advances}.
Also, we omit DRIVE~\cite{vargaftik2021drive} since its performance is identical to that of \name with $\mathcal I_1$ and no packet losses.

Unless otherwise noted, we evaluate the algorithms without variable-length encoding which increases encoding time. We compare with SQGD + Elias Omega encoding~\cite{NIPS2017_6c340f25} and optimized stochastic quantization + Huffman~\cite{pmlr-v70-suresh17a} in Appendix~\ref{app:entropyEval}.

Similarly to Hadamard + SQ, Kashin + SQ, and DRIVE, instead of using a uniform random rotation (which requires $O(d^3)$ time and $O(d^2)$ space) to rotate the vector, we use the randomized Hadamard transform (a.k.a. \emph{structured} random rotation~\cite{pmlr-v70-suresh17a,ailon2009fast}) that admits a fast, \emph{in-place}, parallelizable, and GPU-friendly,  $O(d\log d)$ time implementation~\cite{fino1976unified,uberHadamard,ailon2009fast}. As with these prior works, we find essentially negligible difference in our evaluation between using the Hadamard rotations and fully random rotations. We discuss this further and show supporting empirical measurements in Appendix~\ref{app:hadamard_sr}.

\looseness=-1
\subsection{Implementation Optimization}
In a natural implementation, \name has additional complexity when using $b\neq 1$ bits per coordinate.
Indeed, for $b<1$, we need to sample the sparsification mask, and for $b>1$, we need to identify the interval each rotated coordinate lies in and take its center of mass. The latter can be efficiently done using a binary search - e.g., \texttt{torch.bucketize}, {leading to an encoding complexity of $O(d\cdot b)$).}

Instead, we implement \name using a fine-grained lookup table with a resulting encoding complexity of $O(d)$ (i.e., \emph{independent of $b$}). That is, we map each value $z$ to an integer $n_z=\left\lfloor{\frac{z}{\gamma}}\right\rfloor$ for a suitably selected small value $\gamma$, and our lookup table maps $n_z$ to the message the sender sends.   Similarly, we have a receiver lookup table that maps $n_z$ to an estimated value. The choice of $\gamma$ provides a tradeoff between space and accuracy.
We note that, even with tables whose size is small compared to the encoded vector (e.g., 0.1\%),  the table's granularity is fine enough to get that the additional error is negligible (e.g., less than 0.01\%) compared with the error of the algorithm. Further, it does \emph{not} affect the unbiasedness of the algorithm, which is guaranteed by taking the correct scale (see Theorem~\ref{thm:exentedDriveUnbiasedness}).
This approach allows us to encode and decode coordinates with minimal computation, especially when variable length encoding is not used.

\subsection{$\mathit{vNMSE}$, $\mathit{NMSE}$, and Encoding Speed}\label{sec:prem}

We next evaluate the $\mathit{vNMSE}$ and encoding speed of \name, comparing to three other DME techniques. In Figure \ref{fig:fast_and_accurate_main}, we provide representative results for a bit budget $b=2,4,6$ for vectors that are drawn from a LogNormal(0,1) distribution. Each data point is averaged over 100 trials. 
\name offers the best $\mathit{vNMSE}$ and is faster than Kashin + SQ, which offers the second lowest error. In line with theory, the $\mathit{vNMSE}$ of Hadamard + SQ and the fast QSGD increases with the dimension. In all experiments, \name's encoding time accounts to less than 1\% of the computation of the gradient. Appendix~\ref{app:2f2f} provides further experiments of {$\mathit{NMSE}$ and encode speeds, all indicating similar trends.}

\renewcommand{\thesubfigure}{\roman{subfigure}}

\subsection{Federated Learning}\label{sec:fl-eval}

\begin{figure}[!t]
\hspace*{-1mm}
  \includegraphics[width=1.02\columnwidth]{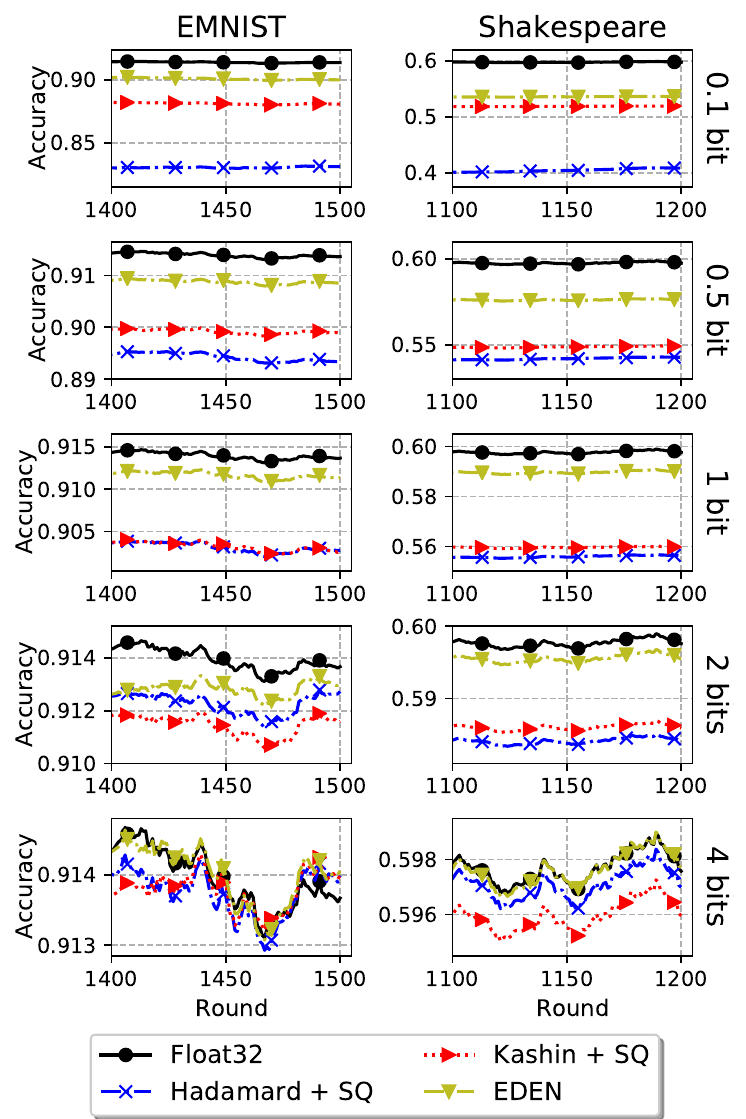}
  \vspace*{-4mm}
  \caption{\emph{FedAvg} over the EMNIST and Shakespeare tasks (columns) at various bit budgets (rows). We report training accuracy per round with a smoothing rolling mean window of 200 rounds. Sparsification is done using a random mask as described in~\S\ref{sec:heteroBits}. Plots are zoomed-in on the last 100 rounds (note the y-axis differences). A zoom-out version is included in Appendix~\ref{app:reddi-expr-details}.
  }
  \centering
  \label{fig:compare_fl}
  \vspace{-4mm}
\end{figure}

We evaluate \name over the federated versions of the \mbox{EMNIST}~\cite{cohen2017emnist} image classification task and the Shakespeare~\cite{shakespeare} next-word prediction task. 
We excluded QSGD, which was less competitive in these experiments.
We run \emph{FedAvg}~\cite{mcmahan2017communication} with the Adam server optimizer~\cite{KingmaB14} and sample $n=10$ clients per round. We re-use code, client partitioning, models, and hyperparameters from the federated learning benchmark of \citet{reddi2021adaptive}. Those are restated for convenience in Appendix~\ref{app:reddi-expr-details}.

Figure~\ref{fig:compare_fl} shows how \name compares with other compression schemes at various bit budgets. We notice that \name considerably outperforms other methods at the lower bit regimes. At 4 bits, all methods converge near the baseline, while \name still maintains a relative advantage.

\subsection{Additional Evaluation} \label{sec:misc-eval}
Due to space limits, we defer additional evaluation results to the Appendix. In particular, we provide experiments for variable-length encoding (Appendix~\ref{app:entropyEval}); structured rotation performance against the theory of uniform rotation (Appendix~\ref{app:hadamard_sr}); $\mathit{NMSE, vNMSE}$, and encoding speed (Appendix~\ref{app:2f2f}); distributed logistic regression (Appendix~\ref{app:lr-eval}); comparison of sub-bit compression and network loss (Appendix~\ref{subsec:losubit}); distributed power iteration (Appendix~\ref{app:pi});  homogeneous federated learning (Appendix~\ref{app:csf}); and cross-device federated learning (Appendix~\ref{app:cdf}). 

\looseness=-1
To summarize these experiments, we show \name outperforms its competitors in nearly all cases, offering a combination of speed, accuracy, overall compression, and robustness, and we believe that this will make it the best choice for many applications.  



\vspace*{-1mm}
\section{Conclusions}

In this paper, we presented \name, a robust and accurate distributed mean estimation technique. \name suits various network scenarios, including packet losses and heterogeneous clients.
Further, we proved strong accuracy guarantees for a wide range of usage scenarios, including using entropy encoding to compress quantized vectors further and working over lossy networks while maintaining high precision.
Our evaluation results indicate that \name considerably outperforms all tested techniques in nearly all settings.

As future work, we propose to study how to combine \name with techniques that provide fast receiver decode procedures, e.g., using a single rotation for all senders to avoid inverse rotating individual vectors. Another direction is to combine \name with techniques such as secure aggregation and differential privacy.
It is also interesting to explore if \name can be adapted to all-reduce techniques, which benefit large-scale distributed deployments where a parameter server might be a bottleneck.
Finally, while \name naturally extends to linear schemes such as weighted mean, we propose to study how to incorporate non-linear aggregation functions, such as approximate geometric median, that may improve the training robustness~\cite{pillutla2022robust}.

Our source code is available at:\\~\url{https://github.com/amitport/EDEN-Distributed-Mean-Estimation}.
\vspace*{-1mm}
\section*{Acknowledgements}
\vspace*{-1mm}

 We thank the anonymous reviews and Moshe Gabel for their insightful feedback and suggestions.  Michael Mitzenmacher was supported in part by NSF grants CCF-2101140, CNS-2107078, and DMS-2023528. Amit Portnoy was supported in part by the Cyber Security Research Center at Ben-Gurion University of the Negev.



\Urlmuskip=0mu plus 1mu\relax
\bibliography{references}
\bibliographystyle{icml2022}


\newpage
\appendix
\onecolumn

\section{Alternative Compression Methods}\label{app:extended_RW}

In this paper, we focus on the DME problem, in which the participants do not keep state, and the estimate of each vector is desired to be unbiased for the $\mathit{NMSE}$ to decrease linearly with respect to the number of senders. We give a few examples of other approaches (i.e., works that do not directly address the DME problem).

Some works (e.g.,~\citet{abs-2002-12410}) investigate the convergence rate of Stochastic Gradient Decent (SGD) for biased compression (which are known to achieve lower error).  
Another approach to leverage the lower error of biased compression is using Error Feedback (EF).
Namely, if the senders are persistent (the same devices are used over multiple rounds) and have the memory to store the error of their compressed gradient, they can use this information to compensate for the estimation error between rounds. Indeed, works such as~\citet{seide20141,alistarh2018convergence,richtarik2021ef21} show that EF-based approaches can greatly increase the accuracy of the learned models and ensure convergence of biased compressors such as Top-$k$ \cite{NEURIPS2018_b440509a} and SketchedSGD~\cite{ivkin2019communication}. 

For a setting with persistent clients, recent works (e.g.,~\citet{mishchenko2019distributed, pmlr-v139-gorbunov21a}) also suggest encoding the difference between the current gradient and the one from the previous round. Intuitively, when the mini-batch sizes are sufficiently large, the sampled gradients are less noisy, and encoding the differences allows faster convergence. This approach is orthogonal to \name which can encode the difference in such a setting.

For distributed cluster learning, some works aim at optimizing streaming aggregation (i.e., All-Reduce operations) via programmable hardware~\cite{sapio2021scaling} or taking advantage of data sparsity~\cite{fei2021efficient}. These approaches are known to be orthogonal to {(and can work in conjunction with) compression techniques~\cite{vargaftik2021drive,fei2021efficient}. For example, if the input is sparse (or is sparsified), one can use \name to encode only the non-zero coordinates.}

Deep gradient compression~\cite{LinHM0D18} leverages redundancy in neural network gradients to reduce the number of transmitted bits. They leverage  momentum correction, local gradient clipping, momentum factor
masking, and warm-up training, and report compression ratios of 270x-600x.

For further overview we refer the reader to~\citet{konecy2017federated,kairouz2019advances,wang2021field}.


\section{\name's Unbiasedness}\label{app:unbiased}

For clarity, we restate the theorem.

\exentedDriveUnbiasedness*

\begin{proof}


Our proof follows similar lines to that of~\citet{vargaftik2021drive}. 
Denote $x'=(\norm x_2,0,\ldots,0)^T$ and let $R_{x\shortrightarrow x'}\in\mathbb R^{d\times d}$ be a rotation matrix such that $R_{x\shortrightarrow x'}\cdot x = x'$.
Further, denote $R_{x} = R R_{x \shortrightarrow x'}^{-1}$. 
Using these definitions we have that,
\begin{equation*}
\begin{aligned}
    \widehat x =& R_{x\shortrightarrow x'}^{-1} \cdot R_{x\shortrightarrow x'}\cdot\widehat x 
    = \s\cdot R_{x\shortrightarrow x'}^{-1} \cdot R_{x\shortrightarrow x'} \cdot   R^{-1}\cdot \mathcal Q\parentheses{\eta_x \cdot R\cdot x}\\ 
    =& \s\cdot R_{x\shortrightarrow x'}^{-1} \cdot  R_x^{-1}\cdot{ \mathcal Q\parentheses{\eta_x \cdot R_x\cdot R_{x\shortrightarrow x'}\cdot x}} 
    = \s\cdot R_{x\shortrightarrow x'}^{-1} \cdot R_x^{-1}\cdot{\mathcal Q\parentheses{\eta_x \cdot R_x\cdot x'}}~. 
\end{aligned}    
\end{equation*}
Let $C_i$ be a vector containing the values of the $i$'th column of $R_x$. Then, $R_x \cdot x'=\norm x_2 \cdot C_0$ and we obtain,
\begin{equation*}
\begin{aligned}
    R_x^{-1}\cdot{ \mathcal Q\parentheses{\eta_x \cdot R_x\cdot x'}} =  \parentheses{\angles{C_0, \mathcal Q \parentheses{\eta_x \cdot \norm{x}_2 \cdot C_0}},\ldots,\angles{C_{d-1},\mathcal Q \parentheses{\eta_x \cdot \norm{x}_2 \cdot C_0}}}^T. 
\end{aligned}   
\end{equation*}
Now, observe that 
$$\angles{R\cdot x,\mathcal Q(\eta_x \cdot R\cdot x)} = \angles{R_x \cdot x',\mathcal Q(\eta_x \cdot R_x \cdot x')} = \norm x_2 \cdot\angles{C_0,\mathcal Q(\eta_x \cdot \norm x_2 \cdot C_0)}~.$$
This yields,
\begin{equation}\label{eq:drive_plus_before_expectation}
\begin{aligned}
&\widehat x = R_{x\shortrightarrow x'}^{-1} \cdot \norm{x}_2 \cdot  
\parentheses{1,\frac{\angles{C_1,\mathcal Q \parentheses{\eta_x \cdot\norm{x}_2 \cdot C_0}}}{\angles{C_0,\mathcal Q \parentheses{\eta_x \cdot\norm{x}_2 \cdot C_0}}},\ldots,\frac{\angles{C_{d-1},\mathcal Q \parentheses{\eta_x \cdot\norm{x}_2 \cdot C_0}}}{\angles{C_0,\mathcal Q \parentheses{\eta_x \cdot\norm{x}_2 \cdot C_0}}}}^T~.
\end{aligned}   
\end{equation} 

Now, consider an algorithm \name' that operates exactly as \name but, instead of directly using the sampled rotation matrix $R = R_x \cdot R_{x \shortrightarrow x'}^{-1}$ it calculates and uses the rotation matrix $R' = R_x \cdot I' \cdot R_{x \shortrightarrow x'}^{-1} = R_{x \shortrightarrow x'} \cdot R \cdot I' \cdot R_{x \shortrightarrow x'}^{-1}$ where $I'$ is identical to the $d$-dimensional identity matrix with the exception that $I'[0,0]=-1$ instead of $1$. 

Since both $R_{x \shortrightarrow x'}$ and $I' \cdot R_{x \shortrightarrow x'}^{-1}$ are fixed rotation matrices, $R'$ and $R$ follow the same distribution.

Now, consider a run of both algorithms where $\widehat x$ is the reconstruction of \name for $x$ with a sampled rotation $R$ and $\widehat x'$ is the corresponding reconstruction of \name' for $x$ with the rotation $R'$. 

According to \eqref{eq:drive_plus_before_expectation} it holds that:
$\widehat x + \widehat x' =R_{x\shortrightarrow x'}^{-1} \cdot \norm{x}_2 \cdot \parentheses{2,0,\ldots,0}^T = 2 \cdot x$. This is because both runs are identical except that the first column of $R_{x}$ and $R_{x} \cdot I'$ have opposite signs and thus for all $i \in \set{1,2,\ldots,d-1}$: 
\begin{equation*}
\begin{aligned}
&\frac{\angles{C_i,\mathcal Q \parentheses{\eta_x \cdot\norm{x}_2 \cdot C_0}}}{\angles{C_0,\mathcal Q \parentheses{\eta_x \cdot\norm{x}_2 \cdot C_0}}} {+~} \frac{\angles{C_i,\mathcal Q \parentheses{\eta_x \cdot\norm{x}_2 \cdot -C_0}}}{\angles{-C_0,\mathcal Q \parentheses{\eta_x \cdot\norm{x}_2 \cdot -C_0}}} {~=~}  \frac{\angles{C_i,\mathcal Q \parentheses{\eta_x \cdot\norm{x}_2 \cdot C_0}}}{\angles{C_0,\mathcal Q \parentheses{\eta_x \cdot\norm{x}_2 \cdot C_0}}} {-~} \frac{\angles{C_i,\mathcal Q \parentheses{\eta_x \cdot\norm{x}_2 \cdot C_0}}}{\angles{C_0,\mathcal Q \parentheses{\eta_x \cdot\norm{x}_2 \cdot C_0}}} {~=~} 0.
\end{aligned}   
\end{equation*} 

Finally, it holds that $\E \brackets{\widehat x + \widehat x'} = 2 \cdot x$. Also, since $R_{x}$ and $R_{x} \cdot I'$ follow the same distribution, due to the linearity of expectation, both algorithms have the same expected value. This yields  $\E \brackets{\widehat x} = \E \brackets{\widehat x'} = x$,
and concludes the proof.
\end{proof}

\section{\name's $\mathit{NMSE}$}\label{app:vnmsetonmselemma}

\vnmsetonmselemma*

\begin{proof}
It holds that, 
\begin{equation*}
\begin{aligned}
\mathit{MSE} =& ~\E\brackets{\norm{\frac{1}{n}\cdot\sum_{c=1}^{n}x_{c}-\frac{1}{n}\cdot\sum_{ c=1}^{n}\widehat{x_{c}}}_2^2}
=\frac{1}{n^2} \cdot \sum_{ c, c'} \E \brackets{\angles{x_{c} - \widehat{x_{c}}, x_{c'} - \widehat{x_{c'}}} } \\  
=& \frac{1}{n^2} \cdot \sum_{ c} \E \brackets{\angles{x_{c} - \widehat{x_{c}}, x_{c} - \widehat{x_{c}}} } + 
\frac{1}{n^2} \cdot \sum_{ c \neq  c'} \E \brackets{\angles{x_{c} - \widehat{x_{c}}, x_{c'} - \widehat{x_{c'}}} } \\
=&\frac{1}{n^2} \sum_{ c} \E \brackets{\norm{x_{c}-\widehat{x_{c}}}_2^2}
=
\frac{1}{n^2} \cdot \sum_{ c} \norm{x_{c}}_2^2 \cdot \E \brackets{\frac{\norm{x_{c}-\widehat{x_{c}}}_2^2}{\norm{x_{c}}_2^2}} 
=
\frac{1}{n^2} \cdot \sum_{ c} \norm{x_{c}}_2^2 \cdot \mathit{vNMSE}(c)~. 
\end{aligned}    
\end{equation*}
Here, we used $\E \brackets{\angles{x_{c} - \widehat{x_{c}}, x_{c'} - \widehat{x_{c'}}} } = 0$. This holds since the estimates of the different clients are unbiased (by Theorem \ref{thm:exentedDriveUnbiasedness}) and independent. Finally, dividing the result by $\frac{1}{n} \cdot \sum_{ c} \norm{x_{c}}_2^2$ yields the result.
\end{proof}

\section{\name's $\mathit{vNMSE}$}\label{app:nmse}

We devide the proof into two parts. First, we prove the main result in \S\ref{subsec:mainr}. Then, for better readability, we defer auxiliary lemmas to \S\ref{subsec:lemmas}.

\subsection{Theorem proof}\label{subsec:mainr}

For clarity, we restate the theorem.

\vnmsetheorem*

\begin{proof}

We begin with bounding the sum of squared errors (SSE). 

The SSE in estimating $\mathcal R(x)$ using $\s\cdot\mathcal Q(\eta_{x}\cdot\mathcal R(x))$ equals that of estimating $x$ using $\widehat x$. Therefore,
\begin{align*}
\norm{x-\widehat x}_2^2 &= \norm{\mathcal R(x-\widehat x)}_2^2 = 
\norm{\mathcal R(x)-\mathcal R(\widehat x)}_2^2 = 
\norm{{\mathcal R(x)}-\mathcal R(\mathcal R^{-1}\big(\s\cdot\mathcal Q(\eta_{x}\cdot\mathcal R(x))\big)\big)}_2^2\\
&=\norm{{\mathcal R(x)}-\mathcal \s\cdot\mathcal Q(\eta_{x}\cdot\mathcal R(x))}_2^2
= \norm{\mathcal R(x)}_2^2 - 2\s \angles{\mathcal R(x), \mathcal Q(\eta_{x}\cdot\mathcal R(x))} + \s^2{\norm{\mathcal Q(\eta_{x}\cdot\mathcal R(x))}_2^2}\\ 
&=\norm{x}_2^2  - 2\s \angles{\mathcal R(x), \mathcal Q(\eta_{x}\cdot\mathcal R(x))} + \s^2{\norm{ \mathcal Q(\eta_{x}\cdot\mathcal R(x))}_2^2}~.    
\end{align*}

Using $S=\frac{\norm{x}_2^2}{\angles{\mathcal R(x), \mathcal Q(\eta_{x}\cdot\mathcal R(x))}}$, the SSE becomes:
\begin{align*}
 \norm{x-\widehat x}_2^2 &= \norm{x}_2^2 
 - 2\s \angles{\mathcal R(x),  \mathcal Q(\eta_{x}\cdot\mathcal R(x))} + \s^2{\norm{  \mathcal Q(\eta_{x}\cdot\mathcal R(x))}_2^2} \\
 &=\frac{\norm{x}_2^4\norm{\mathcal Q(\eta_{x}\cdot\mathcal R(x))}_2^2}{\angles{\mathcal R(x),  \mathcal Q(\eta_{x}\cdot\mathcal R(x))}^2} - \norm{x}_2^2.   
\end{align*}

Thus, the resulting $\mathit{vNMSE}$ in this case is:

$$
\mathbb E\brackets{\frac{\norm{x-\widehat x}_2^2}{\norm{x}_2^2}} = \E \Bigg[\frac{\norm{x}_2^2\norm{\mathcal Q(\eta_{x}\cdot\mathcal R(x))}_2^2}{\angles{\mathcal R(x),  \mathcal Q(\eta_{x}\cdot\mathcal R(x))}^2} \Bigg] - 1~.
$$

Next, since $\mathcal R(x)$ is uniformly distributed on a sphere with a radius $\norm{x}_2$, its distribution is given by $\norm{x}_2 \cdot \frac{Z}{\norm{Z}_2}$ where $Z=(z_1,\ldots,z_d)$ such that $\set{z_i}_{i=1}^d$ are i.i.d random variables and $z_i \sim \mathcal N(0,1) \,\, \forall \, i$. This yields
\begin{align*}
&\frac{\norm{x}_2^2\norm{\mathcal Q(\eta_{x}\cdot\mathcal R(x))}_2^2}{\angles{\mathcal R(x),\mathcal Q(\eta_{x}\cdot\mathcal R(x))}^2} \myeq \frac{\norm{x}_2^2\norm{\mathcal Q(\sqrt d \cdot \frac{Z}{\norm{Z}_2})}_2^2}{\angles{\norm{x}_2\cdot\frac{Z}{\norm{Z}_2},\mathcal Q(\sqrt d \cdot \frac{Z}{\norm{Z}_2})}^2} = \frac{d \cdot \norm{\mathcal Q(\sqrt d \cdot \frac{Z}{\norm{Z}_2})}_2^2}{\angles{\sqrt{d}\cdot\frac{Z}{\norm{Z}_2},\mathcal Q(\sqrt d \cdot \frac{Z}{\norm{Z}_2})}^2} = \frac{d \cdot \norm{\mathcal Q(\Tilde{Z})}_2^2}{\angles{\Tilde{Z},\mathcal Q(\Tilde{Z})}^2}~,  
\end{align*}
where $\myeq$ means equality in distribution and we denote $\Tilde{Z}=\sqrt d \cdot \frac{Z}{\norm{Z}_2}$. 
Thus, our goal is to upper-bound: 
$$ 
\E \sbrac{\frac{d \cdot \norm{\mathcal Q(\Tilde{Z})}_2^2}{\angles{\Tilde{Z},\mathcal Q(\Tilde{Z})}^2}}~. 
$$
For some $0 < \alpha,\beta < \frac{1}{2}$, denote the events 
\begin{align*}
  A=\set{d \cdot (1-\alpha)\leq \norm{Z}_2^2\leq d \cdot (1+\alpha)}~, \quad 
  B=\set{\angles{\tilde{Z},\mathcal Q(\Tilde{Z})} > \frac{\mathbb{E}\Big[ \norm{{\mathcal{Q}}(\Tilde{Z})}_2^2\Big]}{\sqrt{1+\beta}}}~.
\end{align*}
Further denote
$$
f(Z) \triangleq \frac{d \cdot \norm{\mathcal Q(\Tilde{Z})}_2^2}{\angles{\Tilde{Z},\mathcal Q(\Tilde{Z})}^2}~.
$$
Then,
\begin{align*}
f(Z) \le 
&\E\sbrac{ f(Z) \cdot \mathbbm{1}_{A \cap B}} + \sup_Z\para{f(Z)} \cdot \mathbb{P}(A^c\cup B^c)\\ 
\le &\E\sbrac{ f(Z) \cdot \mathbbm{1}_{A \cap B}} + \sup_Z\para{f(Z)} \cdot (\mathbb{P}(A^c) + \mathbb{P}(A\cap B^c))~.
\end{align*}

Next, it holds that
\begin{align*}
\E\sbrac{ f(Z) \cdot \mathbbm{1}_{A \cap B}} \le \E\sbrac{  \frac{d \cdot \norm{\mathcal Q(\Tilde{Z})}_2^2}{\angles{\Tilde{Z},\mathcal Q(\Tilde{Z})}^2} \cdot \mathbbm{1}_{A \cap B}} = 
\E\sbrac{\frac{(1+\beta)\cdot d}{\mathbb{E}\Big[\norm{{\mathcal{Q}}(\Tilde{Z})}_2^2\Big]} \cdot \mathbbm{1}_{A \cap B}} \le \frac{1+\beta}{\mathbb{E}\Big[\para{{\mathcal{Q}}(z)}^2\Big] - \frac{1}{2}\cdot\alpha \cdot M(\mathcal{I})}~.
\end{align*}
In the above, we used Lemma \ref{lem:smaller_ip} by which given that $A$ holds, it holds that $$\mathbb{E}\Big[\norm{{\mathcal{Q}}(\Tilde{Z})}_2^2\Big] \ge \mathbb{E}\sbrac{\norm{{\mathcal{Q}}(\frac{Z}{\sqrt{1+\alpha}})}_2^2} \ge d \cdot \para{ \mathbb{E}\Big[\para{\mathcal{Q}(z)}^2\Big] - (\sqrt{1+\alpha}-1) \cdot M(\mathcal{I})},$$ where $M(\mathcal{I}) > 0$ is a constant that depends on the quantization and we replaced $\sqrt{1+\alpha}-1 \ge \frac{1}{2}\cdot \alpha$ for any $\alpha<\frac{1}{2}$.

Now, we use three Lemmas whose proofs appear in \S\ref{subsec:lemmas}: 
\begin{enumerate}
    \item By Lemma \ref{lem:max_bound} it holds that $\sup_Z\para{f(Z)} \le d^2$.
    \item By Lemma \ref{lem:a} it holds that $\mathbb{P}(A^c) \le 2\cdot e^{-\frac{\alpha^2}{8}\cdot d}$.
    \item By Lemma \ref{lem:b}, for $\alpha = \beta \cdot \frac{0.005}{\sqrt{M(\mathcal{I})}}$ it holds that $\mathbb{P}(A\cap B^c) \le  e^{-\alpha^2\cdot M(\mathcal{I}) \cdot d}$.
\end{enumerate}
These yield 
\begin{align*}
\E \sbrac{f(Z)} \le \frac{1+\alpha\cdot 200 \cdot\sqrt{M(\mathcal{I})}}{ \mathbb{E}\Big[\para{{\mathcal{Q}}(z)}^2\Big] - \frac{1}{2}\cdot\alpha \cdot M(\mathcal{I})}
+ d^2 \cdot \para{2\cdot e^{-\frac{\alpha^2}{8}\cdot d} +e^{-\alpha^2 \cdot M(\mathcal{I})\cdot d}}~,
\end{align*}
where we used $\beta = \alpha \cdot \frac{\sqrt{M(\mathcal{I})}}{0.005} = \alpha\cdot 200 \cdot\sqrt{M(\mathcal{I})}$.
Next, for some constant $k>0$ setting $\alpha=\sqrt{k \cdot \frac{\ln d}{d}}$ yields 
\begin{align*}
\E \sbrac{f(Z)}
\le \frac{1+\sqrt{k \cdot \frac{\ln d}{d}} \cdot 200 \cdot \sqrt{M(\mathcal{I})}}{ \mathbb{E}\Big[\para{{\mathcal{Q}}(z)}^2\Big] - \frac{1}{2}\cdot\sqrt{k \cdot \frac{\ln d}{d}} \cdot M(\mathcal{I})} + 2 \cdot d^2 \cdot \para{ e^{-\frac{k \cdot \ln d}{8}}} +  d^2 \cdot \para{ e^{-k \cdot M(\mathcal{I})\cdot \ln d }}~.
\end{align*}
Let $k=2.5 \cdot\max\set{8,\frac{1}{M(\mathcal{I})}}$. This yields
\begin{align*}
\E \sbrac{f(Z)}
\le \frac{1+\sqrt{\frac{2.5 \cdot\max\set{8,\frac{1}{M(\mathcal{I})}}\ln d}{d}} \cdot 200 \cdot \sqrt{M(\mathcal{I})}}{ \mathbb{E}\Big[\para{{\mathcal{Q}}(z)}^2\Big] - \frac{1}{2}\cdot\sqrt{\frac{2.5 \cdot\max\set{8,\frac{1}{M(\mathcal{I})}}\ln d}{d}} \cdot  M(\mathcal{I})} +\frac{3}{\sqrt d}~~.
\end{align*}
To simplify the asymptotics of the above, we use the lower bound $\mathbb{E}\Big[\para{{\mathcal{Q}}(z)}^2\Big] \ge 0.1$ (from Lemma \ref{app:lemma:qlb}). Thus, for sufficiently large $d$ we find
\begin{align*}
\E \sbrac{f(Z)}
&\le \frac{1+c_1\cdot \sqrt{\frac{\ln d}{d}}}{\mathbb{E}\Big[\para{{\mathcal{Q}}(z)}^2\Big] - c_2\cdot\sqrt{\frac{\ln d}{d}}} + \frac{3}{\sqrt{d}} = \frac{1}{\mathbb{E}\Big[\para{{\mathcal{Q}}(z)}^2\Big] - c_2\cdot\sqrt{\frac{\ln d}{d}}} + \frac{c_1\cdot \sqrt{\frac{\ln d}{d}}}{\mathbb{E}\Big[\para{{\mathcal{Q}}(z)}^2\Big] - c_2\cdot\sqrt{\frac{\ln d}{d}}} + \frac{3}{\sqrt{d}}\\
&= \frac{1 + \frac{c_2\cdot\sqrt{\frac{\ln d}{d}}}{\mathbb{E}\Big[\para{{\mathcal{Q}}(z)}^2\Big] - c_2\cdot\sqrt{\frac{\ln d}{d}}}}{\mathbb{E}\Big[\para{{\mathcal{Q}}(z)}^2\Big]} + \frac{1+c_1\cdot \sqrt{\frac{\ln d}{d}}}{\mathbb{E}\Big[\para{{\mathcal{Q}}(z)}^2\Big] - c_2\cdot\sqrt{\frac{\ln d}{d}}} + \frac{3}{\sqrt{d}}\\
&=\frac{1}{\mathbb{E}\Big[\para{{\mathcal{Q}}(z)}^2\Big]} + \frac{c_2\cdot\sqrt{\frac{\ln d}{d}}}{\mathbb{E}\Big[\para{{\mathcal{Q}}(z)}^2\Big] \cdot \para{\mathbb{E}\Big[\para{{\mathcal{Q}}(z)}^2\Big] - c_2\cdot\sqrt{\frac{\ln d}{d}}}} + \frac{c_1\cdot \sqrt{\frac{\ln d}{d}}}{\mathbb{E}\Big[\para{{\mathcal{Q}}(z)}^2\Big] - c_2\cdot\sqrt{\frac{\ln d}{d}}} + \frac{3}{\sqrt{d}}\\
&\le \frac{1}{\mathbb{E}\Big[\para{{\mathcal{Q}}(z)}^2\Big]} + \frac{c_2\cdot\sqrt{\frac{\ln d}{d}}}{\para{\mathbb{E}\Big[\para{{\mathcal{Q}}(z)}^2\Big]}^2} + \frac{c_1\cdot \sqrt{\frac{\ln d}{d}}}{\mathbb{E}\Big[\para{{\mathcal{Q}}(z)}^2\Big]}  + \frac{3}{\sqrt{d}} = \frac{1}{\mathbb{E}\Big[\para{{\mathcal{Q}}(z)}^2\Big]} + O\para{\sqrt{\frac{\ln d}{d}}}
~~.
\end{align*}
This concludes the proof.
\end{proof}

\subsection{Lemmas proof}\label{subsec:lemmas}

\begin{lemma}\label{lem:max_bound}
It holds that
$$\sup_Z\para{f(Z)} = \sup_Z\para{ \frac{d \cdot \norm{\mathcal Q(\Tilde{Z})}_2^2}{\angles{\Tilde{Z},\mathcal Q(\Tilde{Z})}^2}}\le d^2~.$$
\end{lemma}

\begin{proof}
We can rewrite and obtain
$$
\frac{d \cdot \norm{\mathcal Q(\Tilde{Z})}_2^2}{\angles{\Tilde{Z},\mathcal Q(\Tilde{Z})}^2} = \frac{1}{\langle \frac{Z}{\norm{Z}_2}, \frac{\mathcal Q(\Tilde{Z})}{\norm{\mathcal Q(\Tilde{Z})}_2}  \rangle^2} \le \frac{1}{\para{\frac{1}{d}}^2} = d^2~.
$$
We used the fact that the maximal absolute value entry in a unit vector is at least $\frac{1}{\sqrt{d}}$ and that the maximal entry in both vectors has the same: (1) sign, due the symmetry of the quantization; (2) index, since for any $a_1,a_2 \in \mathbb{R}$ it holds that $a_1\ge a_2 \implies\mathcal Q(a_1)\ge \mathcal Q(a_2)$.
\end{proof}

\begin{lemma}\label{lem:a}
$\mathbb{P}(A^c) \le 2\cdot e^{-\frac{\alpha^2}{8}\cdot d}$~.
\end{lemma}

\begin{proof}
We use a result from~\citet{laurent2000adaptive} (Lemma 1) which we restate here for clarity:\\

\emph{Let U be chosen according to a chi-squared distribution with $D$
degrees of freedom. Then, for any $\lambda>0$: $$\mathbb{P}(U-D \ge 2\sqrt{D\lambda}+2\lambda) \le  e^{-\lambda} \quad \text{and} \quad \mathbb{P}(D-U \ge 2\sqrt{D\lambda}) \le  e^{-\lambda}~.$$}

First, observe the above lemma yields that for any $\lambda>0$:
$$\mathbb{P}(\lvert U-D \rvert \ge 2\sqrt{D\lambda}+2\lambda) \le 2\cdot  e^{-\lambda}~.$$

We want to bound $\mathbb{P}(A^c) = \mathbb{P}\para{ \big\lvert \norm{Z}_2^2 - d \big\rvert \ge \alpha \cdot d}$. First, observe that $\alpha\cdot d \ge 2 \cdot \sqrt{d\cdot\frac{\alpha^2 \cdot d}{8}} + 2\cdot \frac{\alpha^2 \cdot d}{8}$. Next, since $\norm{Z}_2^2$ is chi-squared we obtain,
\begin{align*}
\mathbb{P}(A^c)  =~ &\mathbb{P}\para{\big\lvert \norm{Z}_2^2 - d \big\rvert \ge \alpha \cdot d} \le \\
&\mathbb{P}\para{\big\lvert \norm{Z}_2^2 - d \big\rvert \ge 2\cdot \sqrt{d\cdot\frac{(\alpha^2 \cdot d)}{8}} + 2\cdot \frac{(\alpha^2 \cdot d)}{8}} \le 
2 \cdot  e^{-\frac{(\alpha^2 \cdot d)}{8}}~.\qedhere
\end{align*}
\end{proof}

\begin{lemma}\label{lem:b}
$\mathbb{P}(A\cap B^c) \le  e^{-\alpha^2\cdot M(\mathcal{I}) \cdot d}$ where $M(\mathcal{I})$ is a constant that depends on $\mathcal I$ and $\alpha = \beta \cdot \frac{0.005}{\sqrt{M(\mathcal{I})}}$~.
\end{lemma}
\begin{proof}
Observe that we cannot use a concentration bound directly on $\angles{\tilde{Z},\mathcal Q(\Tilde{Z})}$ since its entries are not independent (i.e., they are normalized by $\norm{Z}_2$). Instead, we rely on event $A$ and use that
\begin{align*}
\mathbb{P}(A \cap B^c) \le~  &\mathbb{P}\para{\angles{\frac{Z}{\sqrt{1+\alpha}},\mathcal Q\para{\frac{Z}{\sqrt{1+\alpha}}}} \le \frac{1}{\sqrt{1+\beta}} \cdot \mathbb{E}\sbrac{\norm{{\mathcal{Q}}\para{\frac{Z}{\sqrt{1-\alpha}}}}_2^2}}~.
\end{align*}

Our goal is to use the following result from~\citet{chung2006concentration} (Theorem 3.5) which we restate here for clarity:\\

\emph{If $X_1, X_2,...,X_n$ are nonnegative independent random variables, we have the following bounds for the sum $X=\sum_{j=1}^n X_j$~: $$\mathbb{P}(X \le \E\sbrac{X}-\lambda) \le  e^{-\frac{\lambda^2}{2\sum_{j=1}^n\E\sbrac{X_j^2}}}~.$$}

To do so, we use that, according to Lemmas \ref{lem:smaller_ip} and \ref{lem:larger_ip},
\begin{itemize}
    \item $\E\sbrac{\angles{\frac{Z}{\sqrt{1+\alpha}},\mathcal Q\para{\frac{Z}{\sqrt{1+\alpha}}}}} \ge \frac{1}{\sqrt{1+\alpha}}\cdot \E\sbrac{\norm{\mathcal Q(Z)}_2^2} - (\sqrt{1+\alpha} - 1)\cdot  M_1(\mathcal{I}) \cdot d$~.
    \item $\E\sbrac{\norm{{\mathcal{Q}}\para{\frac{Z}{\sqrt{1-\alpha}}}}_2^2} \le \frac{1}{\sqrt{1-\alpha}}\E\sbrac{\norm{{\mathcal{Q}}\para{Z}}_2^2} + (1-\sqrt{1-\alpha})\cdot  M_2(\mathcal{I})\cdot d$~.
\end{itemize}
where $M_i(\mathcal{I})$ for $i=1,2$ are finite constants that depend on $\mathcal{I}$.

Next, we have
\begin{align*}
\E&\sbrac{\angles{\frac{Z}{\sqrt{1+\alpha}},\mathcal Q\para{\frac{Z}{\sqrt{1+\alpha}}}}} - \frac{1}{\sqrt{1+\beta}} \cdot  \mathbb{E}\sbrac{\norm{{\mathcal{Q}}\para{\frac{Z}{\sqrt{1-\alpha}}}}_2^2} \\
&\ge\para{\frac{1}{\sqrt{1+\alpha}} \E\sbrac{\norm{\mathcal Q(Z)}_2^2} - (\sqrt{1+\alpha} - 1) M_1(\mathcal{I}) \cdot d} - \frac{1}{\sqrt{1+\beta}}  \para{\frac{1}{\sqrt{1-\alpha}}\E\sbrac{\norm{\mathcal Q(Z)}_2^2} + (1-\sqrt{1-\alpha}) M_2(\mathcal{I}) \cdot d}  \\
&=\E\sbrac{\norm{\mathcal Q(Z)}_2^2} \cdot \para{\frac{1}{\sqrt{1+\alpha}}-\frac{1}{\sqrt{1+\beta}\sqrt{1-\alpha}}} - d \cdot \para{M_2(\mathcal{I})\cdot\frac{(1-\sqrt{1-\alpha})}{\sqrt{1+\beta}} + M_1(\mathcal{I}) \cdot (\sqrt{1+\alpha} -1)}  \\
&=   \para{\frac{0.1}{\sqrt{1+\alpha}} - \frac{0.1}{\sqrt{1+\beta}\sqrt{1-\alpha}}} - d \cdot \para{M_2(\mathcal{I})\cdot\frac{(1-\sqrt{1-\alpha})}{\sqrt{1+\beta}} + M_1(\mathcal{I}) \cdot (\sqrt{1+\alpha}-1)}  \\
&= d \cdot \para{\frac{0.1}{\sqrt{1+\alpha}} - \frac{0.1}{\sqrt{1+\beta}\sqrt{1-\alpha}} -  M_2(\mathcal{I})\cdot\frac{(1-\sqrt{1-\alpha})}{\sqrt{1+\beta}} - M_1(\mathcal{I}) \cdot (\sqrt{1+\alpha}-1)} \triangleq d \cdot \Phi(\alpha, \beta, \mathcal{I})~.
\end{align*}
Here, we used that by Lemma \ref{app:lemma:qlb} it holds that $\E\sbrac{\norm{\mathcal Q(Z)}_2^2} \ge 0.1 \cdot d$ and that our choice of $\alpha,\beta$ results in $\frac{0.1}{\sqrt{1+\alpha}} - \frac{0.1}{\sqrt{1+\beta}\sqrt{1-\alpha}}>0$ (we later show that the constant $M(\mathcal{I})$ is lower bounded by $0.0065$).
Now, we use Taylor expansions around $0$ to simplify $\Phi(\alpha, \beta, \mathcal{I})$. In particular, for $0 \le a \le \frac{1}{2}$: 
\begin{itemize}
    \item $1 + \frac{a}{4} \le \sqrt{1+a} \le 1 + \frac{a}{2}$~.
    \item $1 - a \le \sqrt{1-a} \le 1 - \frac{a}{2}$~.
\end{itemize}

Also, recall that $\alpha,\beta \le \frac{1}{2}$. This yields
\begin{align*}
\Phi(\alpha, \beta, \mathcal{I}) &= \frac{0.1}{\sqrt{1+\alpha}} - \frac{0.1}{\sqrt{1+\beta}\sqrt{1-\alpha}} -  M_2(\mathcal{I})\cdot\frac{(1-\sqrt{1-\alpha})}{\sqrt{1+\beta}} - M_1(\mathcal{I}) \cdot (\sqrt{1+\alpha}-1) \\
& \ge  \frac{0.1}{(1+\frac{\alpha}{2})} - \frac{0.1}{(1+\frac{\beta}{4})(1-\alpha)} - (M_1(\mathcal{I})+M_2(\mathcal{I}))\cdot(\alpha+\frac{\alpha}{2}) \\
&\ge \frac{0.1(1+\frac{\beta}{4})(1-\alpha) - 0.1(1+\frac{\alpha}{2}) - \frac{3}{2}(M_1(\mathcal{I})+M_2(\mathcal{I}))\alpha(1+\frac{\beta}{4})(1-\alpha)(1+\frac{\alpha}{2}) }{(1+\frac{\beta}{4})(1-\alpha)(1+\frac{\alpha}{2})}\\
&\ge \frac{0.025\cdot\beta - \alpha \cdot (0.1625+2.109375(M_1(\mathcal{I})+M_2(\mathcal{I})))}{0.84375}\\
&\ge 0.029\cdot\beta - \alpha\cdot \para{0.2 + 2.5(M_1(\mathcal{I})+M_2(\mathcal{I}))}~.
\end{align*}
Now, if $\alpha\le\frac{0.0145\cdot\beta}{0.2+2.5(M_1(\mathcal{I})+M_2(\mathcal{I}))}$, we have $\Phi(\alpha, \beta, \mathcal{I}) \ge 0.0145\cdot\beta$.
Now we can use the concentration bound and obtain
\begin{align*}
\mathbb{P}(A \cap B^c) &\le~  \mathbb{P}\para{\angles{\frac{Z}{\sqrt{1+\alpha}},\mathcal Q\para{\frac{Z}{\sqrt{1+\alpha}}}} \le \frac{1}{\sqrt{1+\beta}} \cdot \mathbb{E}\sbrac{\norm{{\mathcal{Q}}\para{\frac{Z}{\sqrt{1-\alpha}}}}_2^2}}\\
&\le  e^{-\frac{\para{0.0145\cdot\beta}^2 \cdot d}{6}} =  e^{-M(\mathcal{I})\cdot\alpha^2 \cdot d}~.
\end{align*}
Here, we denoted $M(\mathcal{I}) = \frac{\para{0.2 + 2.5(M_1(\mathcal{I})+M_2(\mathcal{I}))}^2}{6}$ and used that according to Lemma \ref{lem:4mom}, $\E\sbrac{\para{\frac{z}{\sqrt{1+\alpha}} \cdot \mathcal Q\para{\frac{z}{\sqrt{1+\alpha}}}}^2} \le \E\sbrac{\para{z \cdot \mathcal Q\para{z}}^2} = 3$~. Observe that $M(\mathcal{I}) \ge \frac{(0.2)^2}{6} \ge 0.0065$ and thus $\alpha = \beta \cdot \frac{0.005}{\sqrt{M(\mathcal{I})}}$ respects both $\frac{0.1}{\sqrt{1+\alpha}} - \frac{0.1}{\sqrt{1+\beta}\sqrt{1-\alpha}}>0$ and $\alpha\le\frac{0.0145\cdot\beta}{0.2+2.5(M_1(\mathcal{I})+M_2(\mathcal{I}))}$~.
\end{proof}

\newpage
\begin{lemma}\label{lem:smaller_ip}
It holds that,
\begin{itemize}
    \item $\E\sbrac{\angles{\frac{Z}{\sqrt{1+\alpha}},\mathcal Q\para{\frac{Z}{\sqrt{1+\alpha}}}}} \ge \frac{1}{\sqrt{1+\alpha}}\cdot \E\sbrac{\norm{\mathcal Q(Z)}_2^2} - (\sqrt{1+\alpha} - 1)\cdot  M_1(\mathcal{I}) \cdot d$~.
    \item $\mathbb{E}\Big[\norm{{\mathcal{Q}}(\frac{Z}{\sqrt{1+\alpha}})}_2^2\Big] \ge  \E\sbrac{\norm{\mathcal Q(Z)}_2^2} - (\sqrt{1+\alpha} - 1)\cdot  M_1(\mathcal{I}) \cdot d$~.
\end{itemize}
\end{lemma}
\begin{proof}
Due to the linearity of expectation, it is sufficient to show that 
$$
\E\sbrac{\frac{z}{\sqrt{1+\alpha}} \cdot \mathcal{Q}\para{\frac{z}{\sqrt{1+\alpha}}}} \ge \frac{1}{\sqrt{1+\alpha}}\cdot \E\sbrac{\para{\mathcal Q(z)}^2} - (\sqrt{1+\alpha} - 1)\cdot  M_1(\mathcal{I})~,
$$
and
$$
\E\sbrac{\para{\mathcal{Q}\para{\frac{z}{\sqrt{1+\alpha}}}}^2} \ge \E\sbrac{\para{\mathcal Q(z)}^2} - (\sqrt{1+\alpha} - 1)\cdot  M_1(\mathcal{I})~.
$$

Recall the set of intervals $\mathcal{I}$ and denote:
\begin{itemize}
    \item $\mathcal{I}^- = \set{I \in \mathcal{I} | I \subset \mathbb{R}^- \cup \set{0}}$~.
    \item $\mathcal{I}^+ = \set{I \in \mathcal{I} | I \subset \mathbb{R}^+ \cup \set{0}}$~.
\end{itemize}

First, using the law of total expectation, 
\begin{align*}
\E\sbrac{\frac{z}{\sqrt{1+\alpha}} \cdot \mathcal{Q}\para{\frac{z}{\sqrt{1+\alpha}}}} &= \E\sbrac{q_I \cdot \E\sbrac{\frac{z}{\sqrt{1+\alpha}}} ~~\bigg |~~ \frac{z}{\sqrt{1+\alpha}}\in I}  \\
&\ge\frac{1}{\sqrt{1+\alpha}} \sum_{I\in\mathcal{I}} q_I^2 \cdot \mathbb{P}(\frac{z}{\sqrt{1+\alpha}}\in I)  = \frac{1}{\sqrt{1+\alpha}} \cdot \E\sbrac{\para{\mathcal{Q}\para{\frac{z}{\sqrt{1+\alpha}}}}^2}~.    
\end{align*}
Here, we used $\E[z|\frac{z}{\sqrt{1+\alpha}}\in I] \ge \E[z|z\in I] = q_I$.

Next, by definition, $\E\sbrac{\para{\mathcal{Q}\para{\frac{z}{\sqrt{1+\alpha}}}}^2} = \sum_{I\in\mathcal{I}^+} q_I^2 \cdot \mathbb{P}(\frac{z}{\sqrt{1+\alpha}}\in I) + \sum_{I\in\mathcal{I}^-} q_I^2 \cdot \mathbb{P}(\frac{z}{\sqrt{1+\alpha}}\in I)$. Also, since the distribution of $z$ and the set of intervals $\mathcal{I}$ are symmetric around 0,
$\sum_{I\in\mathcal{I}^+} q_I^2 \cdot \mathbb{P}(\frac{z}{\sqrt{1+\alpha}}\in I) = \sum_{I\in\mathcal{I}^-} q_I^2 \cdot \mathbb{P}(\frac{z}{\sqrt{1+\alpha}}\in I)$. For an interval $I\in\mathcal{I}^+$, denote $a_I=\min(I)$. Now, we can write 
$\sum_{I\in\mathcal{I}^+} q_I^2 \cdot \mathbb{P}(\frac{z}{\sqrt{1+\alpha}}\in I) \ge \sum_{I\in\mathcal{I}^+} q_I^2 \cdot \para{\mathbb{P}(z \in I) - \mathbb{P}\para{z \in \sbrac{a_I, a_I\cdot\sqrt{1+\alpha}}}}$. 

Next, we upper-bound $\sum_{I\in\mathcal{I}^+} q_I^2 \cdot \para{\mathbb{P}\para{z \in \sbrac{a_I, a_I\cdot\sqrt{1+\alpha}}}}$. First, 
$
q_I = \frac{\int_{x \in I} x \cdot e^{-\frac{x^2}{2}} dx}{\int_{x \in I} e^{-\frac{x^2}{2}} dx} \le \frac{\int_a^{\infty} x \cdot e^{-\frac{x^2}{2}} dx}{\int_a^{\infty} e^{-\frac{x^2}{2}} dx} \le a_I + \sqrt{\frac{2}{\pi}}~~.
$
Here, the last inequality follows from the fact that the derivative of the hazard rate function of the normal distribution is bounded above by $1$ on the positive reals.
Next, we obtain
\begin{align*}
q_I^2 \cdot \mathbb{P}(z \in \sbrac{a_I, a_I\cdot\sqrt{1+\alpha}}) \le \para{a_I+\sqrt{\frac{2}{\pi}}}^2 \cdot \para{ \frac{1}{\sqrt{2\pi}}\int_{a_I}^{a_I\cdot\sqrt{1+\alpha}} e^{-\frac{x^2}{2}} dx} \le  \frac{\sqrt{1+\alpha}-1}{\sqrt{2\pi}} \cdot a_I \cdot \para{a_I+\sqrt{\frac{2}{\pi}}}^2 \cdot e^{-\frac{a_I^2}{2}}~.
\end{align*}
Thus,
$\sum_{I\in\mathcal{I}^+} q_I^2 \cdot \para{\mathbb{P}\para{z \in \sbrac{a_I, a_I\cdot\sqrt{1+\alpha}}}} \le  \frac{\sqrt{1+\alpha}-1}{\sqrt{2\pi}}\sum_{I\in\mathcal{I}^+} \para{a_I \cdot \para{a_I+\sqrt{\frac{2}{\pi}}}^2 \cdot e^{-\frac{a_I^2}{2}}}$.

Denoting $M_1(\mathcal{I}) = 2 \cdot \frac{1}{\sqrt{2\pi}}\sum_{I\in\mathcal{I}^+} \para{a_I \cdot \para{a_I+\sqrt{\frac{2}{\pi}}}^2 \cdot e^{-\frac{a_I^2}{2}}}$ (we omitted $\sqrt{1+\alpha}$ from the denominator since it only decreases this term) concludes the proof. 

We note that $M_1(\mathcal{I})$ is finite for any $\mathcal{I}$ with a strictly positive and fixed lower bound $\delta_{\mathcal{I}}$ on an interval size. To see this, denote $M = \max_{a\in\mathbb{R}^+}\para{\para{a \cdot \para{a+\sqrt{\frac{2}{\pi}}}^2 \cdot e^{-\frac{a^2}{2}}}}$, ($M\approx 0.589505$) and $a^*$ the corresponding argument ($a^*\approx0.69479$). Then, since for any $a>a^*$ this function is monotonically decreasing,
$$\sum_{I\in\mathcal{I}^+} \para{a_I \cdot \para{a_I+\sqrt{\frac{2}{\pi}}}^2 \cdot e^{-\frac{a_I^2}{2}}} \le \frac{M+\delta_{\mathcal{I}}}{\delta_{\mathcal{I}}} + \sum_{n=0}^\infty (a^* + n\cdot\delta_{\mathcal{I}} + \sqrt{\frac{2}{\pi}})^3\cdot e^{\frac{-(a^* +n\delta_{\mathcal{I}})^2}{2}}~,$$
and the summation converges for any fixed $\delta_{\mathcal{I}}>0$.
\end{proof}

\begin{lemma}\label{lem:larger_ip}
$\E\sbrac{\norm{{\mathcal{Q}}\para{\frac{Z}{\sqrt{1-\alpha}}}}_2^2} \le \frac{1}{\sqrt{1-\alpha}}\E\sbrac{\norm{{\mathcal{Q}}\para{Z}}_2^2} + (1-\sqrt{1-\alpha})\cdot  M_2(\mathcal{I})\cdot d$~.
\end{lemma}

\begin{proof}
Recall the sets of intervals $\mathcal{I}, \mathcal{I}^-, \mathcal{I}^+$ and for an interval $I\in\mathcal{I}^+$, denote $a_I=\min(I)$. 

By definition, $\E\sbrac{\para{\mathcal{Q}(\frac{z}{\sqrt{1-\alpha}})}^2} = \sum_{I\in\mathcal{I}^+} q_I^2 \cdot \mathbb{P}(\frac{z}{\sqrt{1-\alpha}}\in I) + \sum_{I\in\mathcal{I}^-} q_I^2 \cdot \mathbb{P}(\frac{z}{\sqrt{1-\alpha}}\in I)$. Also, since the distribution of $z$ and the set of intervals $\mathcal{I}$ are symmetric around 0, $\sum_{I\in\mathcal{I}^+} q_I^2 \cdot \mathbb{P}(\frac{z}{\sqrt{1-\alpha}}\in I) = \sum_{I\in\mathcal{I}^-} q_I^2 \cdot \mathbb{P}(\frac{z}{\sqrt{1-\alpha}}\in I)$. Now, we can write 
$\sum_{I\in\mathcal{I}^+} q_I^2 \cdot \mathbb{P}(\frac{z}{\sqrt{1-\alpha}}\in I) \le \sum_{I\in\mathcal{I}^+} q_I^2 \cdot \para{\mathbb{P}(z \in I) + \mathbb{P}\para{z \in \sbrac{a_I\cdot\sqrt{1-\alpha}, a_I}}}$. 

Next, we upper-bound $\sum_{I\in\mathcal{I}^+} q_I^2 \cdot \para{\mathbb{P}\para{z \in \sbrac{a_I\cdot\sqrt{1-\alpha}, a_I}}}$. First, we again use $ q_I \le a_I + \sqrt{\frac{2}{\pi}}$ and obtain
\begin{align*}
q_I^2 \cdot \mathbb{P}(z \in \sbrac{ a_I\cdot\sqrt{1-\alpha},a_I}) \le \para{a_I+\sqrt{\frac{2}{\pi}}}^2 \cdot \para{ \frac{1}{\sqrt{2\pi}}\int_{a_I\cdot\sqrt{1-\alpha}}^{a_I} e^{-\frac{x^2}{2}} dx} \le  \frac{1-\sqrt{1-\alpha}}{\sqrt{2\pi}} \cdot a_I \cdot \para{a_I+\sqrt{\frac{2}{\pi}}}^2 \cdot e^{-\frac{a_I^2\cdot(1-\alpha)}{2}}~.
\end{align*}
Thus,
$\sum_{I\in\mathcal{I}^+} q_I^2 \cdot \para{\mathbb{P}\para{z \in \sbrac{ a_I\cdot\sqrt{1-\alpha}, a_I}}} \le  \frac{1-\sqrt{1-\alpha}}{\sqrt{2\pi}}\sum_{I\in\mathcal{I}^+} \para{a_I \cdot \para{a_I+\sqrt{\frac{2}{\pi}}}^2 \cdot e^{-\frac{a_I^2}{4}}}$, where we used $\alpha < \frac{1}{2}$.
Denoting $M_2(\mathcal{I}) = 2 \cdot \frac{1}{\sqrt{2\pi}}\sum_{I\in\mathcal{I}^+} \para{a_I \cdot \para{a_I+\sqrt{\frac{2}{\pi}}}^2 \cdot e^{-\frac{a_I^2}{4}}}$ concludes the proof.

Similarly to the argument in Lemma \ref{lem:smaller_ip}, for any $\mathcal{I}$ with a fixed lower bound on an interval size, $M_2(\mathcal{I})$ is finite.

\end{proof}

\begin{lemma}\label{lem:4mom}
For $z\sim\mathcal{N}(0,1)$ it holds that $\E\sbrac{\para{\frac{z}{\sqrt{1+\alpha}} \cdot \mathcal Q\para{\frac{z}{\sqrt{1+\alpha}}}}^2} \le \E\sbrac{\para{z \cdot \mathcal Q\para{z}}^2} \le 3$.
\end{lemma}
\begin{proof}
It holds that $\para{z^2 - (\mathcal Q(z))^2}^2 = z^4 - 2\cdot z^2 \cdot (\mathcal Q(z))^2  + (\mathcal Q(z))^4 \ge 0$. 

Also, $\E\sbrac{z^2 \cdot (\mathcal Q(z))^2} = \E\sbrac{(\mathcal Q(z))^2 \cdot \E\sbrac{z^2 \mid Q(z)}}$ and by Jensen's inequality and property (v) $\E\sbrac{z^2 \mid Q(z)} \ge \para{\E\sbrac{z \mid Q(z)}}^2 = (Q(z))^2$. Thus, $\E\sbrac{z^2\cdot (\mathcal Q(z))^2}\ge \E\sbrac{(\mathcal Q(z))^4}$. This yields
$$
0< \E\sbrac{\para{z^2 - (\mathcal Q(z))^2}^2} \le \E\sbrac{z^4} - \E\sbrac{(\mathcal Q(z))^4}~.
$$
Next, $$\E\sbrac{(z \cdot \mathcal Q (z))^2} \le  \frac{1}{2}\cdot \E\sbrac{(z)^4}+\frac{1}{2}\cdot\E\sbrac{(\mathcal Q(z))^4} \le \E\sbrac{(z)^4} = 3~.$$
Observing that $z > \frac{z}{\sqrt{1+\alpha}}$ and $\mathcal{Q}(z) \ge \mathcal{Q}\para{\frac{z}{\sqrt{1+\alpha}}}$ for any $\alpha>0$ concludes the proof.
\end{proof}

\begin{lemma}\label{app:lemma:qlb}
$\E\sbrac{\norm{\mathcal Q(Z)}_2^2} \ge 0.1 \cdot d$.
\end{lemma}
\begin{proof}
Due to the linearity of expectation, it is sufficient to show that $\E\sbrac{\para{\mathcal Q(z)}^2} \ge \para{\E\sbrac{\mathcal Q(z)}}^2 \ge 0.1$. 

Also, $\E\sbrac{\mathcal Q(z)} = \sum_{I\in\mathcal{I}} q_I \cdot P(z \in I) = \sum_{I\in\mathcal{I}} \frac{\int_I t \cdot e^{-\frac{t^2}{2}}dt}{\int_I e^{-\frac{t^2}{2}}dt} \cdot \int_I e^{-\frac{t^2}{2}}dt = \sum_{I\in\mathcal{I}} \int_I t \cdot e^{-\frac{t^2}{2}}dt$.

Now,we divide into two cases.

Case 1: $[-a,a] \in \mathcal{I}$. In this case, $a<1$ and thus $\E\sbrac{\para{\mathcal Q(z)}} \ge 2 \cdot \frac{1}{\sqrt{2\pi}}\int_a^\infty e^{\frac{-t^2}{2}dt} > 0.317$.

Case 2: $[-a,a] \not\in \mathcal{I}$ for any $a$. Thus, $\E\sbrac{\para{\mathcal Q(z)}} = 2 \cdot \frac{1}{\sqrt{2\pi}}\int_0^\infty e^{\frac{-t^2}{2}dt} = \sqrt{\frac{2}{\pi}}$.

For both cases, it holds that $\E\sbrac{\para{\mathcal Q(z)}^2} \ge (0.317)^2 \ge 0.1$~.
\end{proof}


\section{Sub-bit Compression Proofs}\label{app:sub-bit}
\smallskip

\twocompressorsvNMSE*
\smallskip

\begin{proof} For ease of exposition we denote $y = \mathcal{A}(x)$ and $z = \mathcal{B}(\mathcal{A}(x))$. Using unbiasedness we obtain:
\begin{align*}
\E\sbrac{\norm{y-z}_2^2 \,|\, y} &= \E\sbrac{\norm{z}_2^2 \,|\, y} - 2\cdot\E\sbrac{\langle z , y \rangle\,|\, y} + \E\sbrac{\norm{y}_2^2}\\
&=\E\sbrac{\norm{z}_2^2 \,|\, y}- \norm{y}_2^2 \implies \E\sbrac{\norm{y-z}_2^2} =\E\sbrac{\norm{z}_2^2} - \E\sbrac{\norm{y}_2^2}~.    
\end{align*}
Similarly, we have that $\E\sbrac{\norm{x-y}_2^2}=\E\sbrac{\norm{y}_2^2} - \norm{x}_2^2$.
Thus, it holds that
$$
\E\sbrac{\norm{x-y}_2^2} + \E\sbrac{\norm{y-z}_2^2} = \E\sbrac{\norm{z}_2^2} - \norm{x}_2^2~.
$$
Also, 
$$
\E\sbrac{\norm{x-z}_2^2} = \E\sbrac{\norm{z}_2^2} - 2\cdot\E\sbrac{\langle x , z \rangle} + \norm{x}_2^2~.
$$
And since, 
$$\E\sbrac{\langle x , z \rangle} = \E\sbrac{\E\sbrac{\langle x , z \rangle} \,|\, y} = \E\sbrac{\langle x , y \rangle} = \norm{x}_2^2~,$$
we obtain,
$$
\E\sbrac{\norm{x-z}_2^2} = \E\sbrac{\norm{z}_2^2} - \norm{x}_2^2 = \E\sbrac{\norm{x-y}_2^2} + \E\sbrac{\norm{y-z}_2^2}~.
$$
\ \\
\textbf{Proof of part 1:\quad} we can write:
\smallskip

\begin{enumerate}
    \item $\frac{\E\sbrac{\norm{x-\mathcal{A}(x)}_2^2}}{\norm{x}_2^2} = \frac{\E\sbrac{\norm{x-y}_2^2}}{\norm{x}_2^2} \le A$~.
    \item $\E\sbrac{\frac{\norm{\mathcal{A}(x)-\mathcal{B}(\mathcal{A}(x))}_2^2}{\norm{\mathcal{A}(x)}_2^2} \,\Bigg|\, \mathcal{A}(x)} = \E\sbrac{\frac{\norm{y-z}_2^2}{\norm{y}_2^2} \,\Bigg|\, y} \le B \implies \E\sbrac{\norm{y-z}_2^2} \le B \cdot \E\sbrac{\norm{y}_2^2}$~.
\end{enumerate}

\smallskip
\vbox{
Using unbiasedness we obtain $\E\sbrac{\norm{x-y}_2^2} = \E\sbrac{\norm{y}_2^2} - \norm{x}_2^2 \le A \cdot \norm{x}_2^2$ and $\norm{x}_2^2 \ge \frac{\E\sbrac{\norm{y}_2^2}}{A+1}$~. Thus, 
\begin{align*}
\frac{\E\sbrac{\norm{x-\mathcal{B}(\mathcal{A}(x))}_2^2}}{\norm{x}_2^2} &= 
\frac{\E\sbrac{\norm{x-z}_2^2}}{\norm{x}_2^2} = \frac{\E\sbrac{\norm{x-y}_2^2}}{\norm{x}_2^2} + \frac{\E\sbrac{\norm{y-z}_2^2}}{\norm{x}_2^2} \\
&\le A + \frac{A+1}{\E\sbrac{\norm{y}_2^2}}\cdot\E\sbrac{\norm{y-z}_2^2} \le A + \frac{A+1}{\E\sbrac{\norm{y}_2^2}}\cdot B \cdot \E\sbrac{\norm{y}_2^2} = A + AB + B~.
\end{align*}
This concludes the proof of part 1.
}

\newpage
\textbf{Proof of part 2:\quad}  The proof is identical to the first part by flipping all the inequalities. We can write:
\begin{enumerate}
    \item $\frac{\E\sbrac{\norm{x-\mathcal{A}(x)}_2^2}}{\norm{x}_2^2} = \frac{\E\sbrac{\norm{x-y}_2^2}}{\norm{x}_2^2} \ge A$~.
    \item $\E\sbrac{\frac{\norm{\mathcal{A}(x)-\mathcal{B}(\mathcal{A}(x))}_2^2}{\norm{\mathcal{A}(x)}_2^2} \,\Bigg|\, \mathcal{A}(x)} = \E\sbrac{\frac{\norm{y-z}_2^2}{\norm{y}_2^2} \,\Bigg|\, y} \ge B \implies \E\sbrac{\norm{y-z}_2^2} \ge B \cdot \E\sbrac{\norm{y}_2^2}$~.
\end{enumerate}
Using unbiasedness we obtain $\E\sbrac{\norm{x-y}_2^2} = \E\sbrac{\norm{y}_2^2} - \norm{x}_2^2 \ge A \cdot \norm{x}_2^2$ and $\norm{x}_2^2 \ge \frac{\E\sbrac{\norm{y}_2^2}}{A+1}$~.
Thus, 
\begin{align*}
\frac{\E\sbrac{\norm{x-\mathcal{B}(\mathcal{A}(x))}_2^2}}{\norm{x}_2^2} &= \E\sbrac{\frac{\norm{x-z}_2^2}{\norm{x}_2^2}} =  \E\sbrac{\frac{\norm{x-y}_2^2}{\norm{x}_2^2}} + \E\sbrac{\frac{\norm{y-z}_2^2}{\norm{x}_2^2}} \\
&\ge A + \frac{A+1}{\E\sbrac{\norm{y}_2^2}}\cdot\E\sbrac{\norm{y-z}_2^2} \ge A + \frac{A+1}{\E\sbrac{\norm{y}_2^2}}\cdot B \cdot \E\sbrac{\norm{y}_2^2} = A + AB + B~.
\end{align*}
This concludes the proof of part 2.
\end{proof}


\section{Lossy Networks Proofs}\label{app:lossy}

\lossylemma*

\begin{proof} The receiver uses $\frac{1}{\norm{m_{ds}}_1}\cdot \mathcal{Q}(\eta_x \cdot R\cdot x) \circ m_{ds}$ instead of $\mathcal{Q}(\eta_x \cdot R\cdot x)$.

\textbf{Proof of 1: \quad} We revisit Equation \eqref{eq:drive_plus_before_expectation} in Theorem \ref{thm:exentedDriveUnbiasedness} and obtain
\begin{align*}
\widehat x = R_{x\shortrightarrow x'}^{-1} \cdot \norm{x}_2 \cdot\parentheses{\frac{\angles{C_0,\frac{1}{\norm{m_{ds}}_1}\cdot \mathcal Q \parentheses{\eta_x \cdot\norm{x}_2 \cdot C_0}\circ m_{ds}}}{\angles{C_0,\mathcal Q \parentheses{\eta_x \cdot\norm{x}_2 \cdot C_0}}},\ldots,\frac{\angles{C_{d-1},\frac{1}{\norm{m_{ds}}_1}\cdot \mathcal Q \parentheses{\eta_x \cdot\norm{x}_2 \cdot C_0}\circ m_{ds}}}{\angles{C_0,\mathcal Q \parentheses{\eta_x \cdot\norm{x}_2 \cdot C_0}}}}^T~.
\end{align*} 

The proof continues similarly to that of Theorem \ref{thm:exentedDriveUnbiasedness}.

We again consider algorithm \name' from the proof of Theorem \ref{thm:exentedDriveUnbiasedness}. It holds that:
$$\widehat x + \widehat x' = 2 \cdot R_{x\shortrightarrow x'}^{-1} \cdot \norm{x}_2 \cdot \parentheses{\frac{\angles{C_0,\frac{1}{\norm{m_{ds}}_1}\cdot \mathcal Q \parentheses{\eta_x \cdot\norm{x}_2 \cdot C_0}\circ m_{ds}}}{\angles{C_0,\mathcal Q \parentheses{\eta_x \cdot\norm{x}_2 \cdot C_0}}},0,\ldots,0}^T~.$$
In the numerator, we have a sum of random variables whose all subsets of size $\norm{m_{ds}}_1$ follows the same distribution. This means that for any two deterministic masks $m_{ds}$ and $m_{ds}^{'}$ such that $\norm{m_{ds}}_1 = \norm{m_{ds}^{'}}_1$, we have that
\begin{align*}
\E\sbrac{\frac{\angles{C_0,\frac{1}{\norm{m_{ds}}_1}\cdot \mathcal Q \parentheses{\eta_x \cdot\norm{x}_2 \cdot C_0}\circ m_{ds}}}{\angles{C_0,\mathcal Q \parentheses{\eta_x \cdot\norm{x}_2 \cdot C_0}}}}= \E\sbrac{\frac{\angles{C_0,\frac{1}{|m_{ds}^{'}|}\cdot \mathcal Q \parentheses{\eta_x \cdot\norm{x}_2 \cdot C_0}\circ m_{ds}^{'}}}{\angles{C_0,\mathcal Q \parentheses{\eta_x \cdot\norm{x}_2 \cdot C_0}}}} ~.
\end{align*}
Also, due to linearity of expectation,
\begin{align*}
\sum_{\substack{m_{ds}':\\ \norm{m_{ds}'}_1=\norm{m_{ds}}_1}}& \E\sbrac{\frac{\angles{C_0,\frac{1}{p}\cdot \mathcal Q \parentheses{\eta_x \cdot\norm{x}_2 \cdot C_0}\circ m_{ds}'}}{\angles{C_0,\mathcal Q \parentheses{\eta_x \cdot\norm{x}_2 \cdot C_0}}}} = \\
&\frac{1}{p}\cdot \E\sbrac{\sum_{\substack{m_{ds}':\\ \norm{m_{ds}'}_1=\norm{m_{ds}}_1}} \frac{\angles{C_0,\mathcal Q \parentheses{\eta_x \cdot\norm{x}_2 \cdot C_0}\circ m_{ds}'}}{\angles{C_0,\mathcal Q \parentheses{\eta_x \cdot\norm{x}_2 \cdot C_0}}}} = \frac{1}{p}\cdot  p \cdot M_{ds}~,
\end{align*}
where $M_{ds}={d \choose \norm{m_{ds}}_1}$ is the number of different masks with the same number of 1's. This means that $\E \brackets{\widehat x + \widehat x'} = 2 \cdot x$. Recall that $\widehat x$ and $\widehat x'$ follow the same distribution. This concludes the proof.

Intuitively, the reason why unbiasedness is preserved with a deterministic mask is that all coordinates of the rotated and quantized vector follow the same distribution, and thus after the inverse rotation and scaling, the distribution of the reconstructed vector depends only on the number of zeros in the mask but not on their indices. 

\textbf{Proof of 2: \quad} Due to unbiasedness (i.e., Part 1), it holds that
$$
\mathit{vNMSE} \cdot \norm{x}_2^2 = \E\sbrac{\norm{x-\widehat x}_2^2} =  \norm{\widehat x}_2^2 -  \norm{x}_2^2~.
$$
Now, we examine $\norm{\widehat x}_2^2 = \norm{\mathcal {R}(\widehat x)}_2^2 = \frac{1}{\norm{m_{ds}}_1^2} \cdot \norm{S\cdot \mathcal{Q}(\eta_x \cdot R\cdot x) \circ m_{ds}}_2^2$.

Similarly to Case 1, we have a sum of random variables whose all subsets of size $\norm{m_{ds}}_1$ follows the same distribution. This means that for any two deterministic masks $m_{ds}$ and $m_{ds}^{'}$ such that $\norm{m_{ds}}_1 = \norm{m_{ds}^{'}}_1$, we obtain $\E\sbrac{\frac{1}{\norm{m_{ds}}_1^2} \cdot \norm{S\cdot \mathcal{Q}(\eta_x \cdot R\cdot x) \circ m_{ds}}_2^2} = \E\sbrac{\frac{1}{\norm{m_{ds}'}_1^2} \cdot \norm{S\cdot \mathcal{Q}(\eta_x \cdot R\cdot x) \circ m_{ds}'}_2^2}$~.

Let $m_{rs}$ be a random mask such that $\norm{m_{rs}}_1 = \norm{m_{ds}}_1$. Then,
\begin{multline*}
\E\sbrac{\frac{1}{\norm{m_{rs}}_1^2} \cdot \norm{S\cdot \mathcal{Q}(\eta_x \cdot R\cdot x) \circ m_{rs}}_2^2}\\
= \frac{1}{M_{ds}}\cdot \sum_{\substack{m_{ds}':\\ \norm{m_{ds}'}_1=\norm{m_{ds}}_1}} \E\sbrac{\frac{1}{\norm{m_{ds}'}_1^2} \cdot \norm{S\cdot \mathcal{Q}(\eta_x \cdot R\cdot x) \circ m_{ds}'}_2^2 \,\Bigg|\, m_{rs}=m_{ds}'} \\
 = \E\sbrac{\frac{1}{\norm{m_{ds}}_1^2} \cdot \norm{S\cdot \mathcal{Q}(\eta_x \cdot R\cdot x) \circ m_{ds}}_2^2}
~.   \qquad\qquad\qquad
\end{multline*}
Using the above with Lemma~\ref{lemma:twocompressorsvNMSE} concludes the proof.
\end{proof}


\begin{figure}[h!]
\centering
  \includegraphics[width=0.75\textwidth]{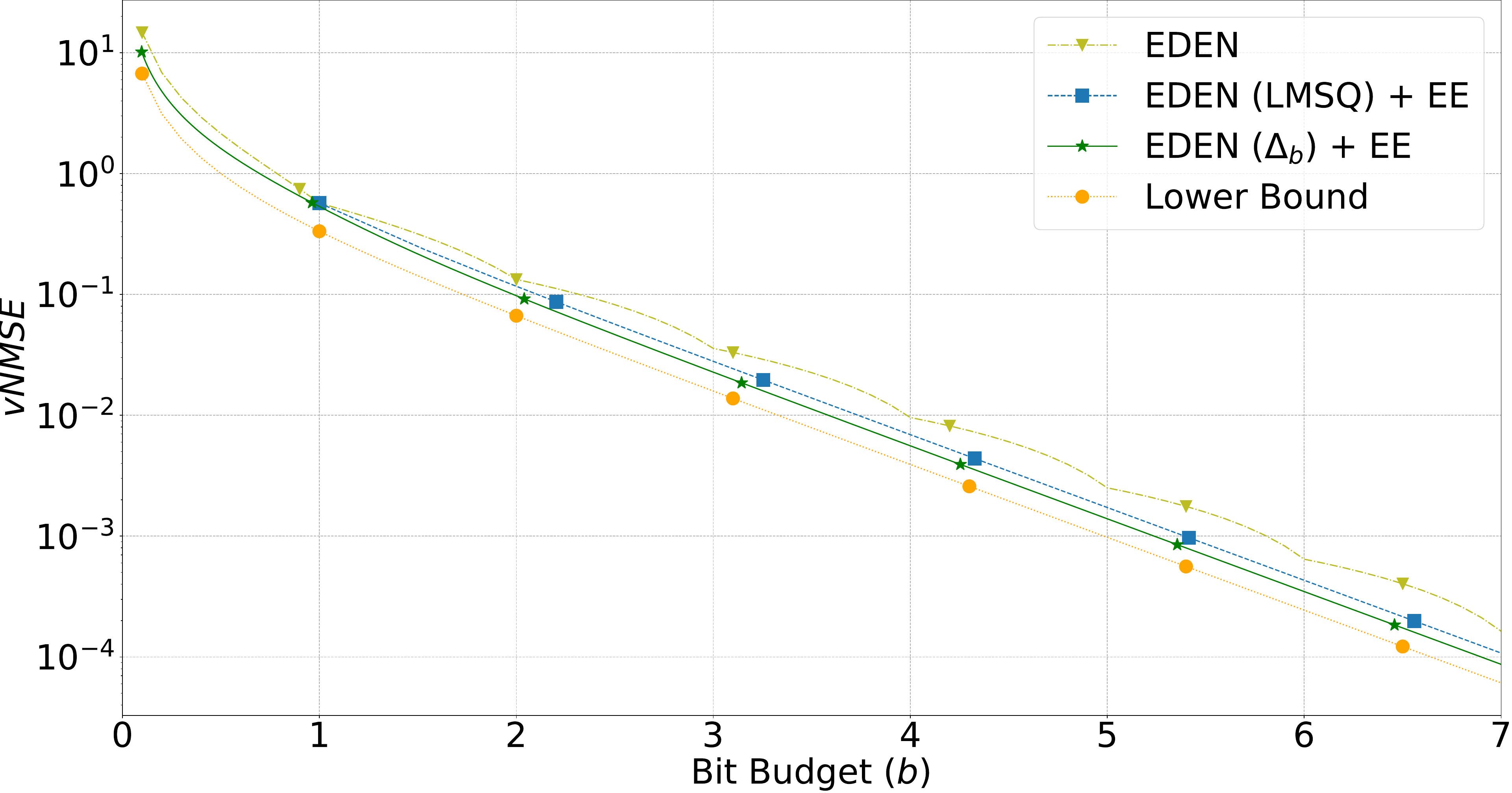}
  \caption{\name's $\mathit{vNMSE}$ with and without Entropy Encoding for the different quantization schemes. (displayed for $b\in\set{0.1\cdot i \mid i\in \set{1,\ldots,80}}$)}
  \label{fig:EDEN_vNMSE_EE}
\end{figure}
\section{Entropy Compressed \name}
\subsection{Evaluation}\label{app:entropyEval}

As discussed in~\S\ref{sec:entropy}, when using Entropy Encoding (EE), \name uses the quantization interval set $\mathcal I_{\Delta_b}=\set{\brackets{\Delta_b\cdot\para{n-\frac{1}{2}},\Delta_b\cdot\para{n+\frac{1}{2}}} \Big | n\in\mathbb Z}$, for the smallest $\Delta_b$ such that $H_{\mathcal{I}_{\Delta_b}}\le b$.

\mbox{Figure~\ref{fig:EDEN_vNMSE_EE} shows the $\mathit{vNMSE}$ of:}
\begin{itemize}
    \item \name with the super-bit compression (\S\ref{sec:heteroBits}) for $b\ge 1$ and the sub-bit compression (\S\ref{sec:heteroBits}) for $b\in(0,1]$.
    \item \name, with EE applied on the vector resulting from the Lloyd-Max Scalar Quantizer.
    \item \name, with EE applied on the vector resulting from the $\mathcal Q_{\mathcal I_{\Delta_b}}$ quantization.
    \item A lower bound on \name, for any $\mathcal I$, derived from the Rate-distortion theory~\cite{cover1999elements} {over the normal distribution.}
\end{itemize}

As shown, even without EE, \name requires less than one bit more than the lower bound.
Indeed, using more quantization values and compressing the resulting vector with EE reduces the error. Switching to our tailored quantization $\mathcal I_{\Delta_b}$ reduces the error further. Also, $\mathcal I_{\Delta_b}$ requires at most $0.25$ bits more than the rate-distortion lower bound.

Next, we compare \name, with and without EE, to two previously suggested variable-length encoding DME schemes. 
Specifically, we show the $\mathit{vNMSE}$ of QSGD with Elias Omega encoding~\cite{NIPS2017_6c340f25}, and optimized stocastic quantization with Huffman encoding~\cite{pmlr-v70-suresh17a}.
Without EE, \name uses the Lloyd Max Scalar Quantizer $\mathcal I_b$ for $b\ge 1$ (\S\ref{sec:oq}) while for $b\in(0,1]$ it uses the sub-bit compression (\S\ref{sec:heteroBits}).
When EE is applied, \name uses  $\mathcal I_{\Delta_b}$ (see \S\ref{sec:entropy}).

\begin{figure}[h]
\centering
  \includegraphics[width=0.99\textwidth]{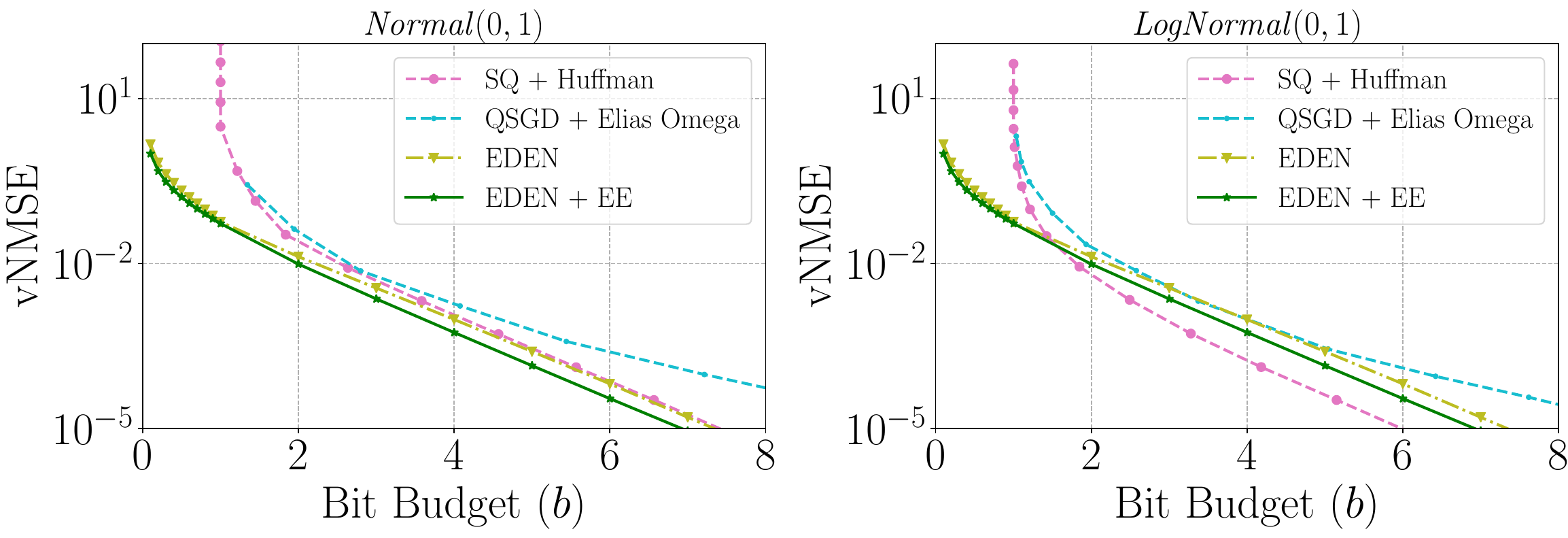}
  \vspace*{-2mm}
  \caption{\name's $\mathit{vNMSE}$ with and without Entropy Encoding compared to other Variable-Length schemes. (displayed for $b\in\set{0.1,0.2,0.3,0.4,0.5,0.6,0.7,0.8,0.9,1,2,3,4,5,6,7,8}$)}
  \vspace*{-2mm}
  \label{fig:ee_cs_ee}
\end{figure}

As depicted in Figure \ref{fig:ee_cs_ee}, \name, which has a GPU-friendly implementation (i.e., without EE), improves the worst-case $\mathit{vNMSE}$. Also, \name is robust and has the same error for all distributions (which is aligned with the theoretical results) and can be further optimized using EE, at the cost of additional computation. The stochastic quantization with Huffman encoding has a lower $\mathit{vNMSE}$ for some input distributions, e.g., the LogNormal, but its error is input dependent.


\subsection{Variable-Length Encoding Representation Length}\label{app:entropy}

In practice, the frequency of a quantization value $I\in\mathcal I$ may not be exactly $p_I\cdot d$.
Nonetheless, using arithmetic-coding, we can get an encoding that uses $(H_{\mathcal I}+\epsilon)(1+o(1))$ bits per coordinate on average.
A proof sketch, which assumes that the coordinates are independent (in practice, they are weakly dependent for a sufficiently large dimension $d$) follows. We defer the formal proof to future work.
As indicated by~\citet{mitzenmacher2017probability} (see Chapter 10), from the fact that each coordinate $j$ of a vector $y\in\mathbb R^d$, decreases the length of the encoded interval by a factor of $\mathcal L_{\mathcal Q(y[j])}$, where $\mathcal L_{\mathcal Q(y[j])}$ is the random variable that represents the quantization value of the interval of $y[j]$.
Therefore, the length of the interval of the vector is $\mathcal L_y\triangleq\prod_{j=0}^{d-1}\mathcal L_{\mathcal Q(y[j])}$ which means that the representation of $y$ requires $\ceil{\log_2{(\frac{1}{\mathcal L_y})}+1}\le 2+\sum_{j=0}^{d-1}\log_2 \frac{1}{\mathcal L_{\mathcal Q(y[j])}}$. A standard application of the Chernoff bound suffices to complete the argument.


\section{Structured Rotation}\label{app:hadamard_sr}

To improve computational efficiency, the randomized Hadamard transform is used in different domains to replace computationally extensive matrix multiplications. This is now common.  For example: in ~\citet{rader1969new,thomas2013parallel,herendi1997fast}, it is used to develop a computationally cheap methods to generate independent normally distributed variables from simpler (e.g., uniform) distributions; in \citet{yu2016orthogonal}, it is used in the context of Gaussian kernel approximation, replacing the random Gaussian matrix; in \citet{choromanski2018structured}, it is used for gradient estimation in derivative-free
optimization reinforcement learning; in \citet{choromanski2017unreasonable}, it is used for efficient  computation of embeddings.

In our context, it is used by recent DME techniques (Hadamard + SQ~\cite{pmlr-v70-suresh17a,konevcny2018randomized}, Kashin + SQ~\cite{lyubarskii2010uncertainty,caldas2018expanding,safaryan2020uncertainty} and DRIVE~\cite{vargaftik2021drive}). 
While~\cite{vargaftik2021drive} pointed out an adversarial example for a single transform, we are not aware of adversarial inputs for more than a single transform. 
For some use cases, previous works (e.g.,~\citet{yu2016orthogonal,10.5555/2969239.2969376}) suggest using 2-3 transforms to avert the dependency on the input.
In the case of neural networks, as was indicated by~\cite{vargaftik2021drive}, when the input vector is high dimensional and admits finite moments, a single transform performs sufficiently similar to a uniform rotation. As recently reported by several works, this is indeed the case in common DNN workloads where gradients {and network parameters follow, e.g., the lognormal~\cite{chmiel2020neural} or normal \cite{banner2018post,ye2020accelerating} distributions.}

\begin{figure}[h]
\centering
  \includegraphics[width=0.99\textwidth]{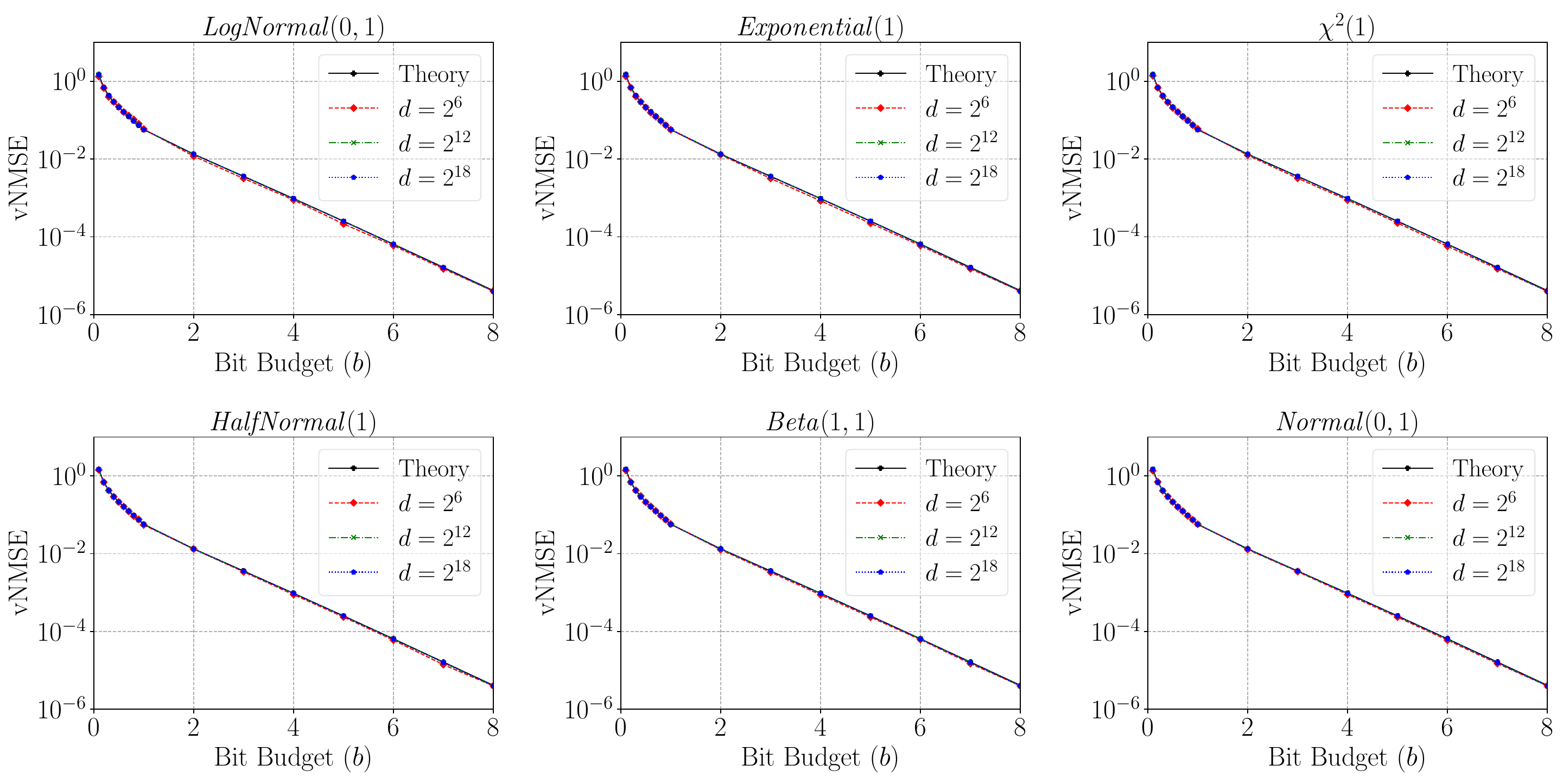}
  \vspace*{-4mm}
  \caption{Using a single Hadamard transform (instead of a uniform random rotation) results in a $\mathit{vNMSE}$ that coinsides with the theoretical bound of Corollary \ref{corr:nmse} for these tested distributions (displayed for $b\in\set{0.1,0.2,0.3,0.4,0.5,0.6,0.7,0.8,0.9,1,2,3,4,5,6,7,8}$).}
  \label{fig:hadamard_is_good}
  \vspace*{-0mm}
\end{figure}

In Figure~\ref{fig:hadamard_is_good}, we show how \name's $\mathit{vNMSE}$ with a single transform aligns with the theoretical  bound (Corollary~\ref{corr:nmse}) for all tested input distributions (LogNormal, Normal, Exponential, $\chi^2$, Half-Normal, Beta, and others not shown) and vector dimensions. As can be seen, even a single randomized Hadamard transform aligns extremely closely with the theoretical results for a uniform random rotation, so that one cannot visually distinguish the results.  For even smaller dimensions (e.g., $d=16$), we observed a slightly higher error for the implementation than the theoretical asymptotic bound, as the $O\parentheses{\sqrt{\frac{\log d}{d}}}$ term of Theorem~\ref{thm:nmse} is not negligible. However, since our interest is in neural network gradients of large dimension, this is not important for this application.


\section{Additional Simulation Details and Results}\label{app:sims}

\subsection{$\mathit{vNMSE}$, $\mathit{NMSE}$, and encoding speed}\label{app:2f2f}

\begin{figure}[t]
     \centering
     \subfloat[$b=1$ bit budget.]{
         \includegraphics[width=0.5\textwidth]{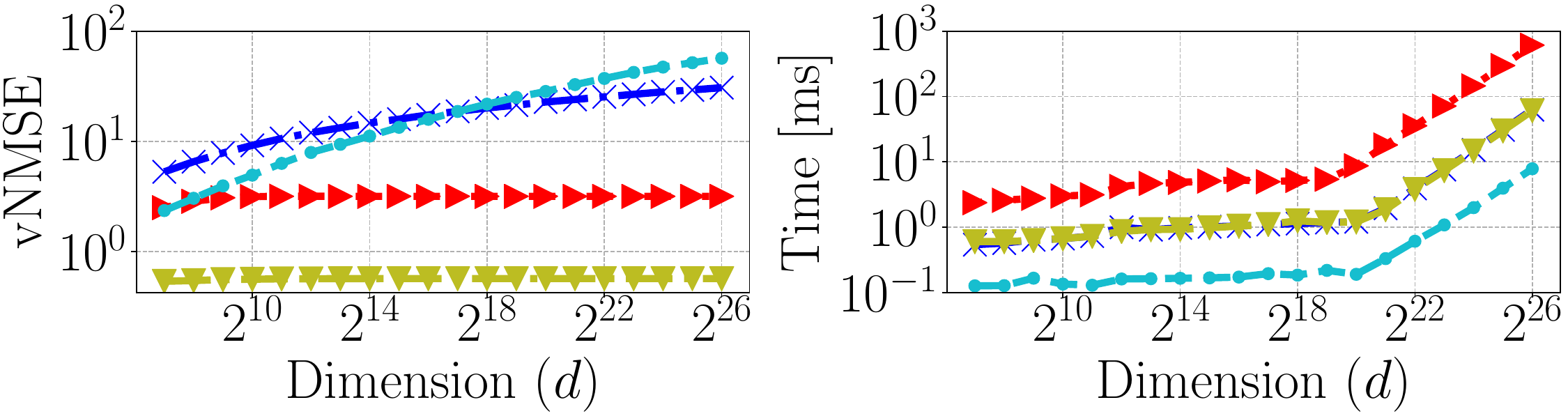}
         \label{fig:nmse:normal1}
     }
     \subfloat[$b=3$ bit budget.]{
         \includegraphics[width=0.5\textwidth]{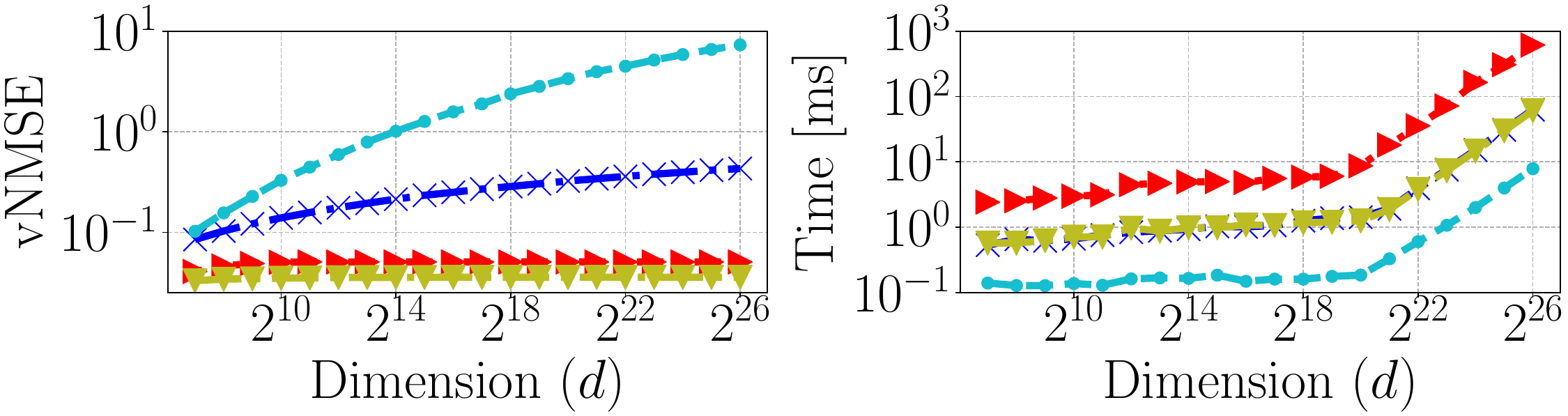}
         \label{fig:nmse:normal3}
     }     \\
     \subfloat[$b=5$ bit budget.]{
         \includegraphics[width=0.5\textwidth]{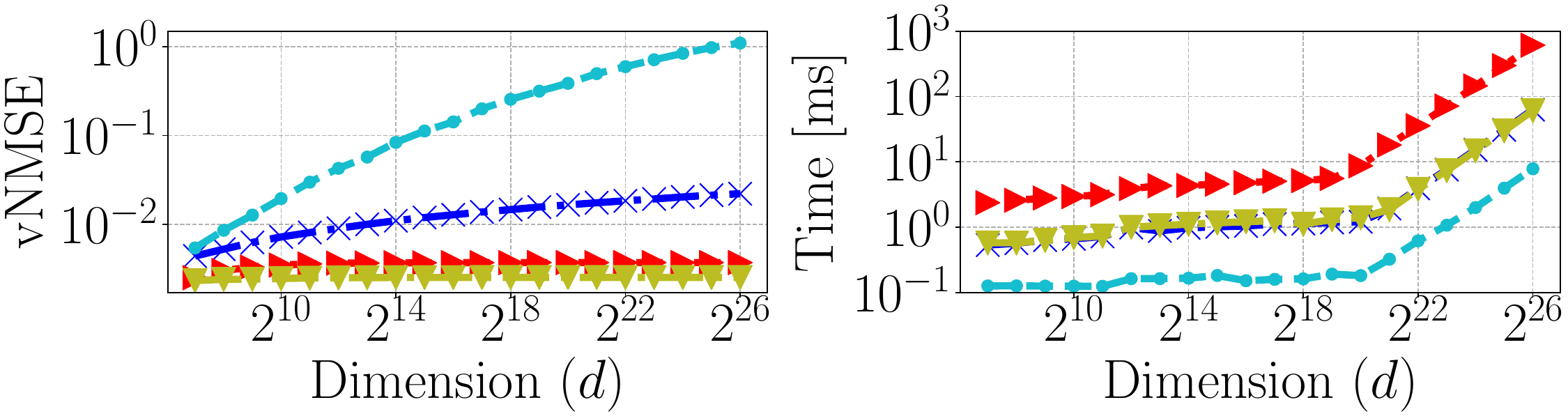}
         \label{fig:nmse:normal5}
     }
     \subfloat[$b=7$ bit budget.]{
         \includegraphics[width=0.5\textwidth]{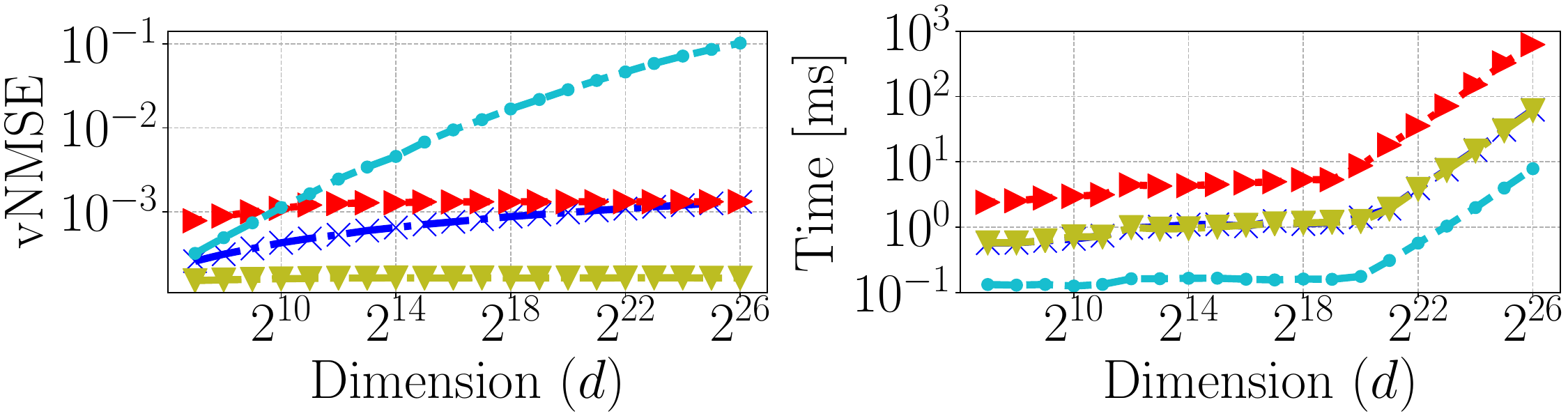}
         \label{fig:nmse:normal7}
     } \\
     \vspace{-2mm}
     \subfloat{
         \centering
         \includegraphics[width=0.7\textwidth]{figures/encoding_legend-cropped.pdf}
         \label{fig:nmse:normalLegend}
     }
     \vspace{-2mm}
     \caption{The $\mathit{vNMSE}$ and compression time as a function of the dimension $d$ for LogNormal(0,1) distribution.}
     \label{fig:fast_and_accurate_357}
\end{figure}

Here, we run the experiment of Figure~\ref{fig:fast_and_accurate_main} with different bit budgets $b$ and depict the results in Figure~\ref{fig:fast_and_accurate_357}.
As shown, \name has the lowest $\mathit{vNMSE}$ for all dimension and bit budget combinations, and is also significantly faster than the second most accurate solution, Kashin + SQ. As in Figure~\ref{fig:fast_and_accurate_main}, the $\mathit{vNMSE}$ of \name and Kashin + SQ is bounded independently of the dimension while Hadamard + SQ's and QSGD's error increases with $d$. 

Our encoding speed measurements are performed using NVIDIA GeForce RTX 3090 GPU. The machine has Intel Core i9-10980XE CPU (18 cores, 3.00 GHz, and 24.75 MB cache) and 128 GB RAM. We use Ubuntu 20.04.2 LTS operating system, CUDA release 11.1 (V11.1.105), and PyTorch version 1.10.1. 

\begin{figure}[t]
 \centering
    \includegraphics[width=\textwidth]{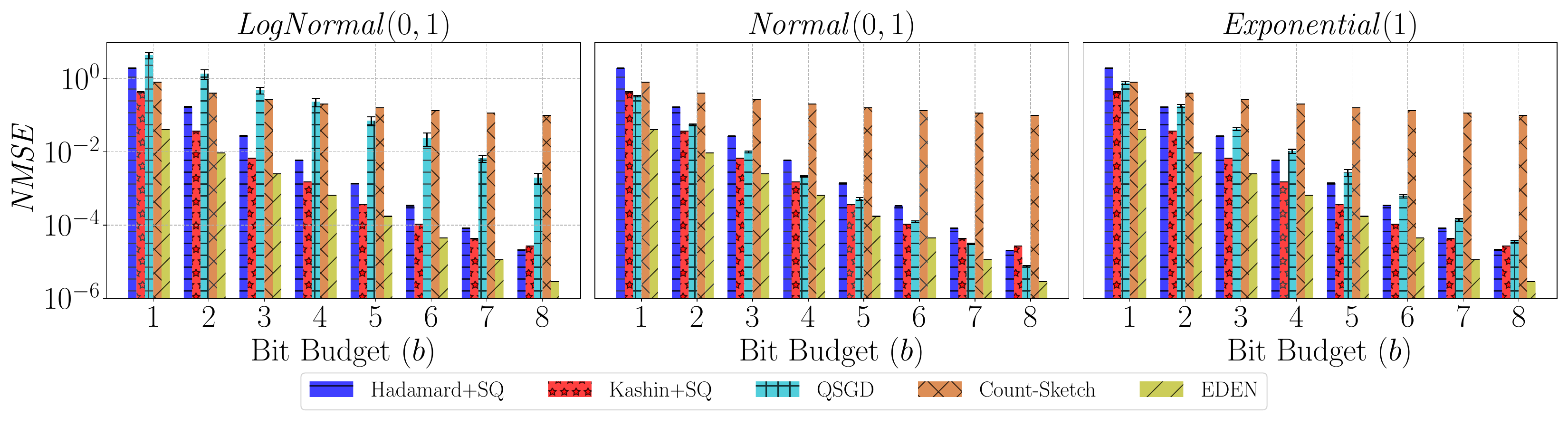}
     \caption{$\mathit{NMSE}$ evaluation with $c=10$ clients and vector dimension of $d=11,511,784$ (ResNet18). Sweeping over a bit budget of 1-8 bits per coordinate.}
     \vspace{-3mm}
     \label{fig:nmse:three_dists}
\end{figure}

Next, we conduct experiments with a vector of size $d=11511784$ (the number of parameters in a ResNet18 architecture) in a DME setting where $c=10$ senders send their vectors to a receiver for averaging. We test the different DME techniques over different distributions. Each data point is repeated 100 times and we report the mean and standard deviation. In addition, here we also compare against \emph{Count-Sketch}~\cite{charikar2002finding}, which is the main building block for some recent distributed and federated compression schemes (e.g.,~\citet{ivkin2019communication}).

{
We sweep over a bit budget of 1-8 bits per coordinate. The results, shown in Figure~\ref{fig:nmse:three_dists}, indicate that: (1) as dictated by theory, \name's $\mathit{NMSE}$ is not sensitive to the specific distribution; (2) \name significantly improves over all techniques for all bit budgets (notice the logarithmic $\mathit{NMSE}$ scale). Moreover, there is no consistent runner-up in this DME experiment since different competitors are more accurate for different distributions and bit budgets. 
}

Several additional trends are evident: (1) Count-Sketch is competitive for a low communication budget but not for a higher ones. This is expected since its error guarantee decreases polynomially in $b$ (which linearly affects its number of counters). In contrast, for other techniques, the error decreases exponentially (i.e., doubling the number of quantization values for each added bit); (2) For most bit budgets, the most competitive DME algorithm to ours in terms of $\mathit{NMSE}$ is Kashin + SQ. However, its accuracy is less prominent for high bit budgets constraints where its coefficient representation dominates the error (which can be improved with more iterations, at the cost of additional computation); (3) QSGD employs uniform quantization values and therefore performs better for light tailed distributions. For example, for the heavy-tailed Log-Normal distribution, which is common in machine learning workloads and, in particular, neural network gradients (e.g.,~\citet{chmiel2020neural}), its performance degrades notably. 

Finally, recall that for $b=1$ and no coordinate losses, \name and DRIVE~\cite{vargaftik2021drive} admit the same performance.

\subsection{EMNIST and Shakespeare experiments details}\label{app:reddi-expr-details}

The EMNIST and Shakespeare federated learning experiments, presented in~\S\ref{sec:fl-eval}, follow precisely the setup described in~\citet{reddi2021adaptive} for the Adam server optimizer case, which we restate in Table~\ref{table:reddi_params}.

The client partitioning of these datasets was designed to naturally simulate a realistic heterogeneous federated learning setting. Specifically, federated EMNIST~\cite{caldas2019leaf} includes 749,068 handwritten characters partitioned among their 3400 writers (i.e., this is the total number of clients), and federated Shakespeare~\cite{mcmahan2017communication} consists of 18,424 lines of text from Shakespeare plays partitioned among the respective 715 speakers (i.e., clients). For EMNIST, a CNN with two convolutional layers is used (with ${\approx}1.2M$ parameters), and for Shakespeare, a standard LSTM recurrent model \cite{Hochreiter1997LongSM} (with ${\approx}820K$ parameters).

\begin{table*}[h]
\centering
\begin{tabular}{|l||c|c|c|c|c|c|}
\hline
\textbf{Task} & \textbf{Clients per round} & \textbf{Rounds} & \textbf{Batch size} & \textbf{Client lr} & \textbf{Server lr} & \textbf{Adam's $\epsilon$} \\ \hline \hline
EMNIST & 10   & 1500  & 20 & $10^{-1.5}$  & $10^{-2.5}$ & $10^{-4}$ \\ \hline
Shakespeare & 10 & 1200 & 4 & 1  &  $10^{-2}$ & $10^{-3}$ \\ \hline
\end{tabular}
\caption{Hyperparameters for the EMNIST and Shakespeare experiments.}
\label{table:reddi_params} 
\end{table*}

\begin{figure}[h]
\centering
  \includegraphics[width=\textwidth]{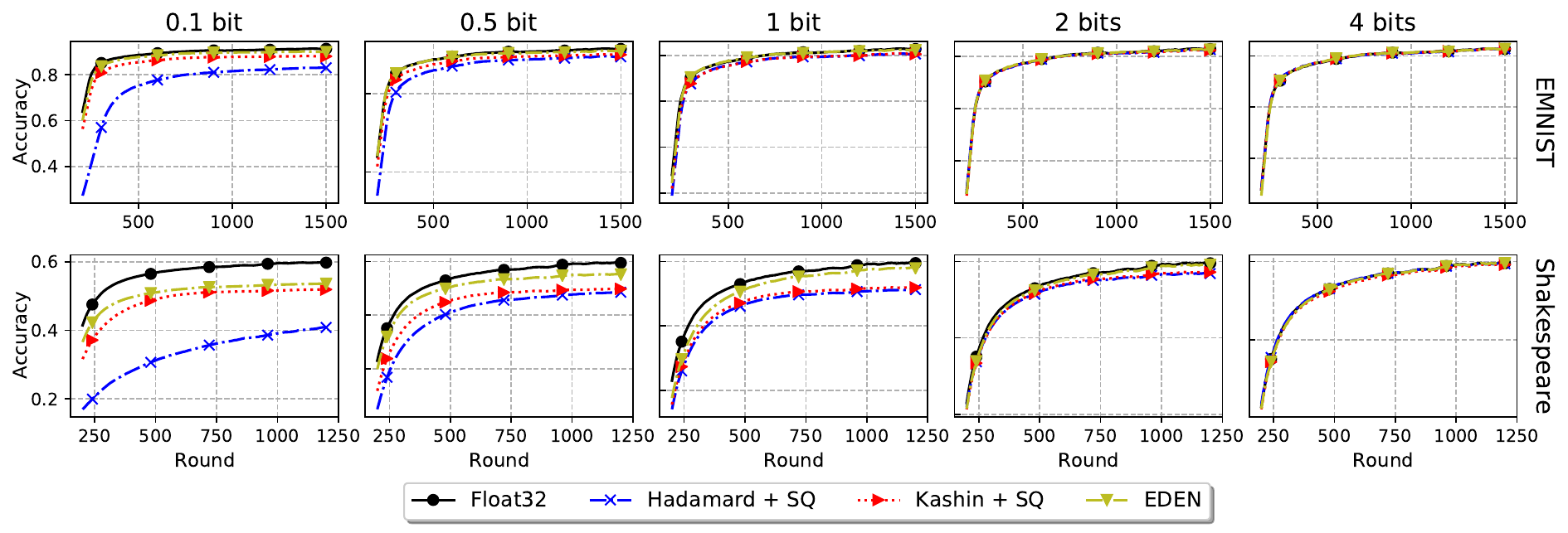}
  \caption{A fully-zoomed out of Figure~\ref{fig:compare_fl}. \emph{FedAvg} over the EMNIST and Shakespeare tasks (columns) at various bit budgets (rows). We report training accuracy per round with a smoothing rolling mean window of 200 rounds. Sparsification is done using a random mask as described in \S\ref{sec:heteroBits}.
  }
  \label{fig:compare_fl_no_zoom}
\end{figure}

Additionally, for completeness, we include Figure~\ref{fig:compare_fl_no_zoom}, which is a zoomed-out version of Figure~\ref{fig:compare_fl}.

\subsection{Distributed Logistic Regression} \label{app:lr-eval}

\begin{figure}[h]
  \begin{center}
  \includegraphics[width=.52\textwidth]{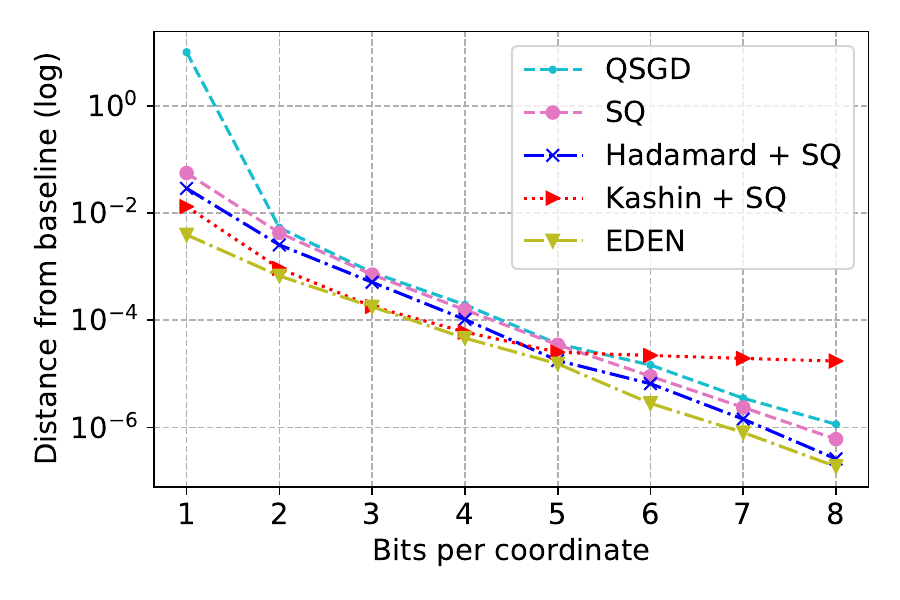}
  \end{center}
  \caption{Logistic regression over UCI’s Census Income task. For every compression scheme and bit budget, we run 100 rounds of distributed gradient descent and report the Euclidean distance from the baseline model (i.e., Float32).}
  \centering
  \label{fig:logistic_regression}
\end{figure}

While the federated learning benchmarks demonstrate the applicability of our method, they are also often noisy and generally converge with low bit budgets. In contrast, logistic regression allows us to show more fine-grained differences between the methods for super-bit budgets. We perform an experiment similar to that of \citet{DBLP:conf/icml/MalinovskiyKGCR20} (\S 4). In particular, we use UCI's Census Income binary prediction task~\cite{kohavi1996scaling} with the discretization described in~\cite{platt1998fast}\footnote{Available at \url{https://www.csie.ntu.edu.tw/~cjlin/libsvmtools/datasets/binary.html\#a9a}.} and divide the data equally among 20 clients. With each compression strategy, we run distributed gradient descent where, in each round, clients report the full gradient over their share, and the server uses the average of these gradients to take a step towards the optimum.  Given that this is a convex setup, we expect the lower variance unbiased gradient estimates to consistently move towards the optimum. We run each compression scheme for 100 rounds with different bit budgets and report the Euclidean distance of the model parameters from the model received without any compression. We present the results in Figure~\ref{fig:logistic_regression}. As hypothesized, \name is closer to the optimum than other methods in all the bit budgets we measured.

\subsection{Loss vs. Sub-bit}\label{subsec:losubit}

\begin{figure}[h]
\centering
  \includegraphics[width=0.6\textwidth]{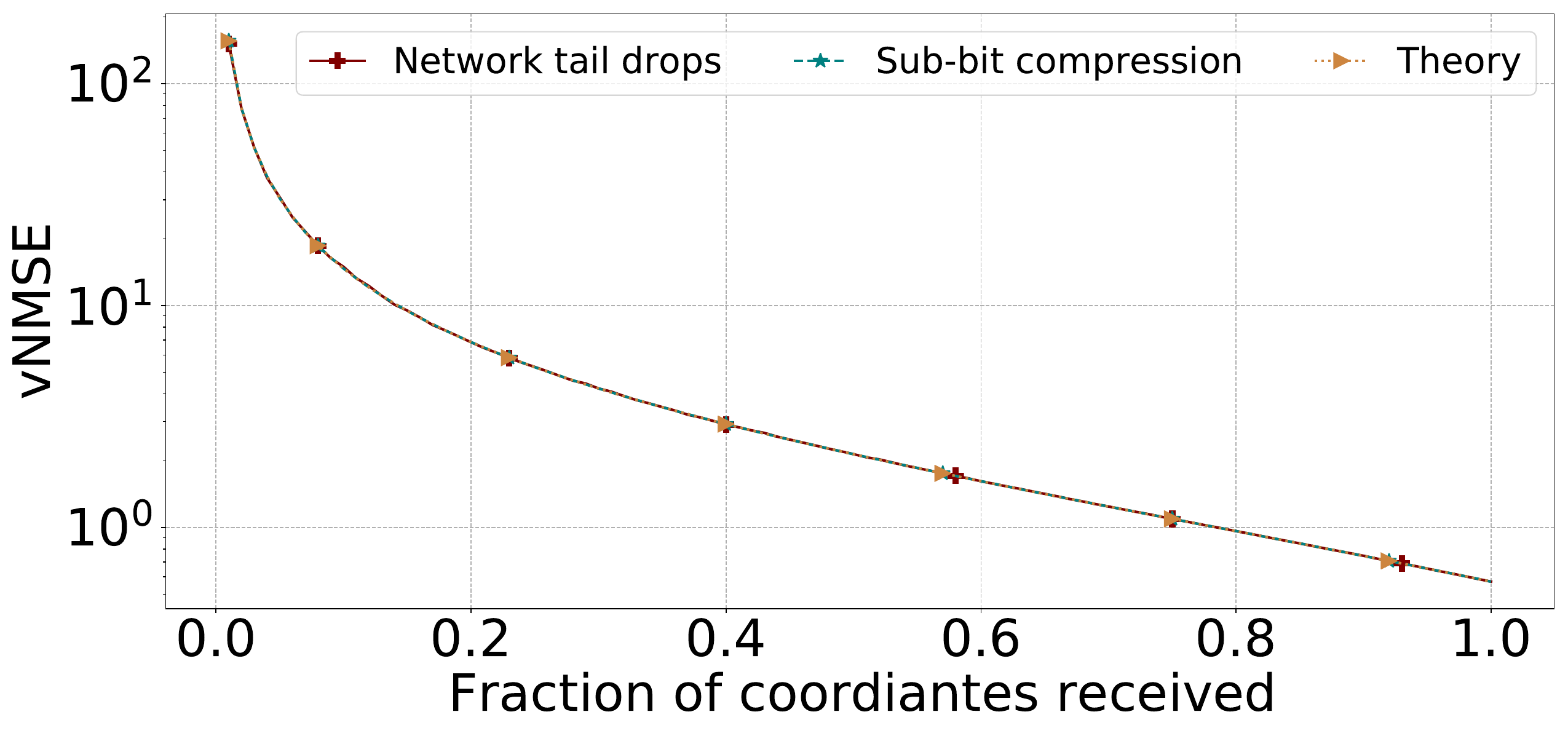}
  \caption{\name with sub-bit compression vs. network tail drops vs. the theoretical bound ($d=2^{19}$).}
  \label{fig:losubit}
\end{figure}

We now measure the $\mathit{vNMSE}$ under network loss and for sub-bit compression.
Intuitively, while in both not all coordinates are received by the receiver, our sub-bit compression selects a uniform random mask of coordinates that are encoded while packet loss may be arbitrary (e.g., in blocks).
A different view point is that sub-bit compression means sparsifying the vector \emph{prior} to the random rotation while packet drops means loss of coordinates in the rotated vector.
As shown in Figure~\ref{fig:losubit}, the empirical $\mathit{vNMSE}$ of the two is identical (when the same fraction of coordinates is received) and follows the theory of Corollary~\ref{corr:subitsim} and Lemma~\ref{lemma:lossylemma}.
Indeed, as shown by Lemma~\ref{lemma:twocompressorsvNMSE}, the $\mathit{vNMSE}$ equals $A+AB+B$ (where $A$ is \name's $\mathit{vNMSE}$ and B is the sparsification's) regardless of the orders in which $\mathcal A$ and $\mathcal B$ are applied.

\subsection{Distributed Power Iteration}\label{app:pi}

\begin{figure}
  \centering
  \includegraphics[trim={0 1.8cm 0 0},clip,width=0.68\textwidth]{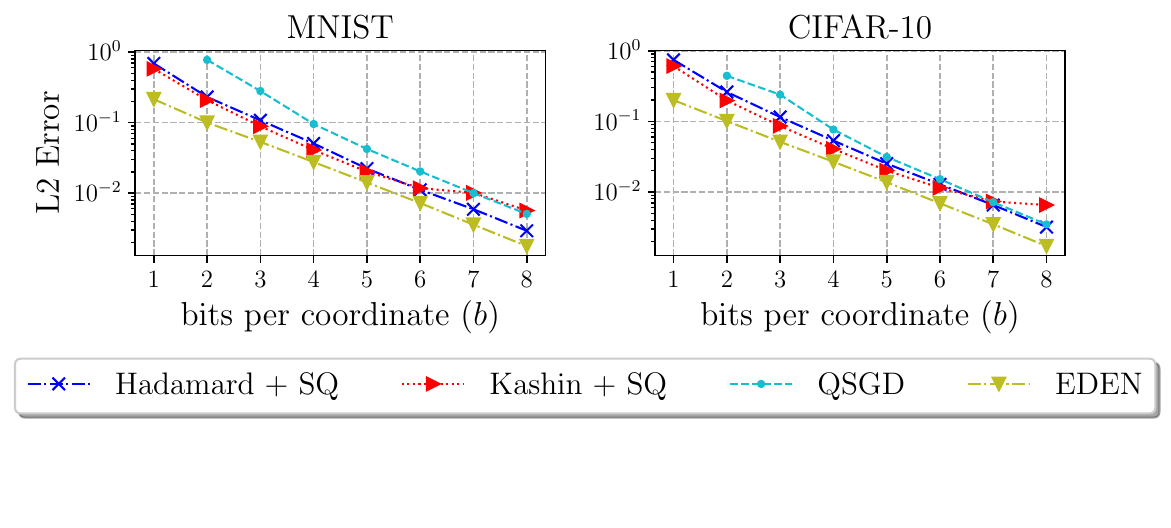}
  \caption{Distributed power iteration of MNIST and CIFAR-10 with 10 clients.}
  \label{fig:power_iteration}
\end{figure}

For some machine learning tasks (e.g., Principal Component Analysis), power iteration, which approximates the dominant eigenvalues and eigenvectors of a matrix, is used as a sub-routine. We perform an experiment with $10$ clients and a server that jointly compute the top eigenvector in a matrix that is distributed among the clients. In particular, the training occurs in rounds where in each training round we have the following sequence of events:
\begin{itemize}
    \item Each client updates the top eigenvector based on its local data, compresses it, and sends it to the server. 
    \item For each client, the server calculates the diff vector between the eigenvector from the previous round, averages the diffs and updates the eigenvector estimation using this average scaled by a \emph{learning rate} of 0.1.  
    \item The server sends the updated eigenvector estimation to all of the clients.
\end{itemize}

For each compression scheme, we vary the bit budget $b$ from one bit to eight and measure the L2 norm of the diff between the final eigenvector to the optimal solution (i.e., the achieved eigenvector without compression).
Figure~\ref{fig:power_iteration} presents the results for the MNIST and CIFAR-10~\cite{krizhevsky2009learning, lecun1998gradient, lecun2010mnist} datasets.
It is evident that in both datasets, \name achieves the lowest L2 error for all values of $b$.

\subsection{Homogeneous federated learning}\label{app:csf}


\begin{figure}[h!]
  \centering
  \includegraphics[trim={0 2.6cm 0 0},clip,width=0.98\textwidth]{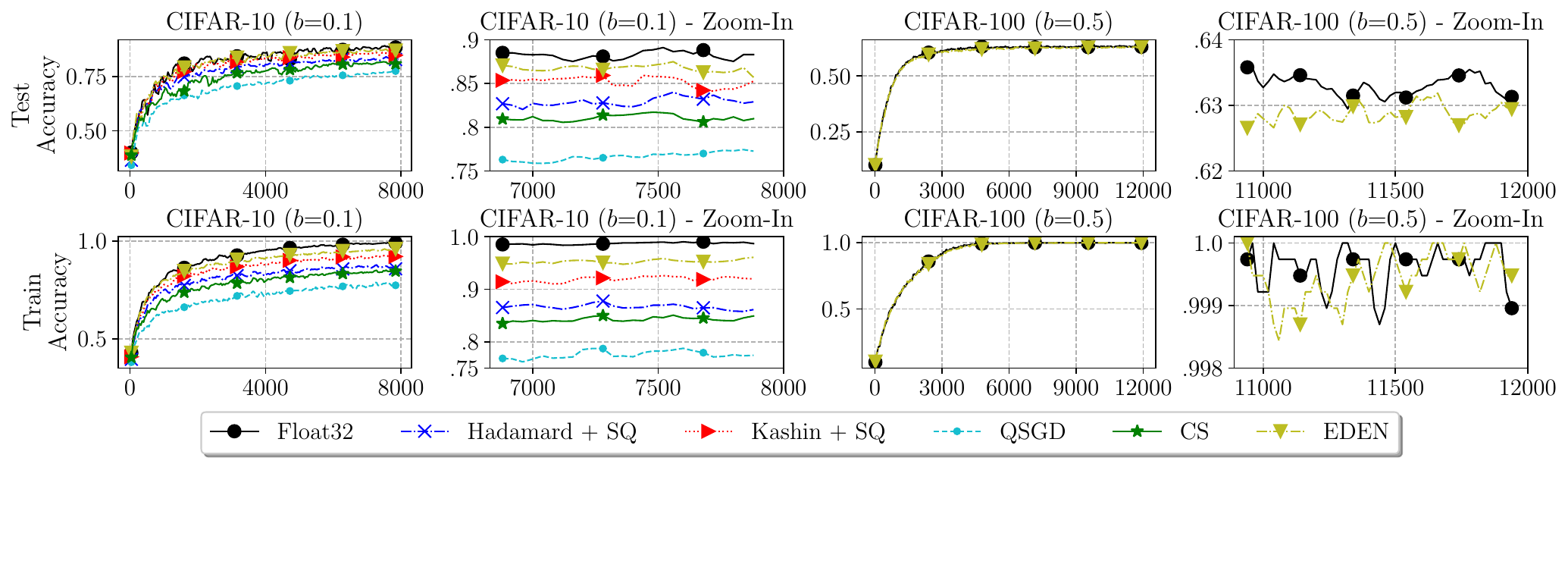}
  \caption{Homogeneous federated learning of CIFAR-10 and CIFAR-100 with 10 clients.}
  \label{fig:cross_silo}
\end{figure}

We simulate two federated scenarios with 10 clients and the CIFAR-10 and CIFAR-100 datasets~\cite{krizhevsky2009learning}. For both scenarios the data is uniformly distributed among the clients. We use a batch size of 128, an SGD optimizer with a Cross entropy loss criterion, and each client performs a single training step at each round. 
For CIFAR-10, we use the ResNet-9~\cite{he2016deep} architecture with learning rate of 0.1, and the bit budget $b$ is set to 0.1.
For CIFAR-100, we use the ResNet-18~\cite{he2016deep} architecture with learning rate of 0.05, and the bit budget $b$ is set to 0.5.

Figure~\ref{fig:cross_silo} presents the test and train accuracy of \name and competitive compression schemes, with a smoothing rolling
mean window of 60 rounds. In both scenarios, \name achieves the highest accuracy. More specifically, for CIFAR-100, \name is the only compression scheme that converges for such a low bit budget.

\subsection{Cross-device federated learning}\label{app:cdf}

\begin{figure}
  \centering
  \includegraphics[trim={0 2.6cm 0 0},clip,width=0.98\textwidth]{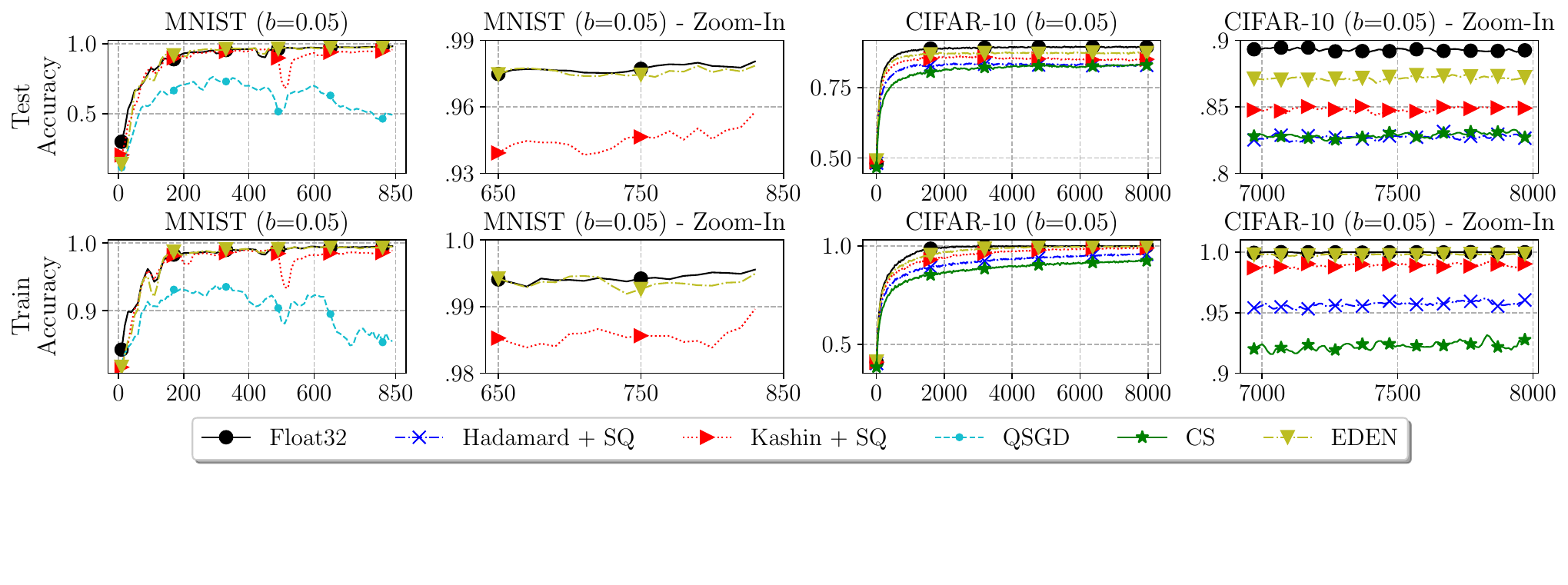}
  \caption{Cross-device federated learning of MNIST and CIFAR-10 with 50 clients.}
  \label{fig:cross_device}
\end{figure}

We simulate two cross-device federated scenarios with 50 clients and the MNIST and CIFAR-10 datasets~\cite{krizhevsky2009learning, lecun1998gradient, lecun2010mnist}. For both scenarios, 10 clients are randomly chosen at each training round, and each client performs five local training steps. 
We use a batch size of 128, an SGD optimizer with a cross entropy loss criterion, and the bit budget $b$ is set to 0.05, testing a setting with very aggressive compression.

For MNIST, we use LeNet-5~\cite{lecun1998gradient} with a learning rate of 0.05. 
To simulate severe heterogeneity, each client holds data instances that belong to a single class (i.e., severe label skew).
For the CIFAR-10 configuration, we use ResNet-9~\cite{he2016deep} with a learning rate of 0.1, and the data is uniformly distributed among the clients.

Figure~\ref{fig:cross_device} presents the test and train accuracy of \name and the competitive compression schemes, with a smoothing rolling
mean window of 30 rounds. In both scenarios, \name achieves the highest accuracy. For MNIST, \name is only slightly below the baseline model (without compression).

\end{document}